\documentclass[review]{elsarticle}
\usepackage{graphicx}
\usepackage{comment}
\usepackage{amsmath,amssymb} 
\usepackage{color}
\usepackage{url}            
\usepackage{booktabs}       
\usepackage{amsfonts}       
\usepackage{nicefrac}       
\usepackage{microtype}      
\usepackage{enumerate}
\usepackage[american]{babel}
\usepackage{graphicx}
\graphicspath{{figure/}}
\usepackage{subfig}
\usepackage{threeparttable}
\usepackage{amssymb}
\usepackage{multirow}
\usepackage{marvosym}
\usepackage{cancel}

\usepackage{float}

\def\oo{\phantom{0}}
\journal{Pattern Recognition}

\begin{document}

\begin{frontmatter}

\title{Neural Random Subspace}

\author[mymainaddress]{Yun-Hao Cao}
\ead{caoyh@lamda.nju.edu.cn}

\author[mymainaddress]{Jianxin Wu\corref{mycorrespondingauthor}}
\cortext[mycorrespondingauthor]{Corresponding author}
\ead{wujx2001@nju.edu.cn}

\author[mysecondaryaddress]{Hanchen Wang}
\ead{hw501@cam.ac.uk}

\author[mysecondaryaddress]{Joan Lasenby}
\ead{jl221@cam.ac.uk}

\address[mymainaddress]{State Key Laboratory for Novel Software Technology, Nanjing University, China}
\address[mysecondaryaddress]{Department of Engineering, University of Cambridge, UK}

\begin{abstract}
The random subspace method, known as the pillar of random forests, is good at making precise and robust predictions. However, there is not a straightforward way yet to combine it with deep learning.
In this paper, we therefore propose Neural Random Subspace (NRS), a novel deep learning based random subspace method. 
In contrast to previous forest methods, 
NRS enjoys the benefits of end-to-end, data-driven representation learning, as well as pervasive support from deep learning software and hardware platforms, hence achieving faster inference speed and higher accuracy. 
Furthermore, as a non-linear component to be encoded into Convolutional Neural Networks (CNNs), NRS learns non-linear feature representations in CNNs more efficiently than previous higher-order pooling methods, producing good results with negligible increase in parameters, floating point operations (FLOPs) and real running time. Compared with random subspaces, random forests and gradient boosting decision trees (GBDTs), NRS achieves superior performance on 35 machine learning datasets. Moreover, on both 2D image and 3D point cloud recognition tasks,
integration of NRS with CNN architectures achieves consistent improvements 
with minor extra cost.
\end{abstract}

\begin{keyword}
random subspace, ensemble learning, deep neural networks
\end{keyword}

\end{frontmatter}


\section{Introduction}
Deep convolutional neural networks (CNNs) have achieved remarkable advancements in a variety of computer vision tasks~\cite{advances18:PR}, such as image classification~\cite{ vgg:simonyan:ICLR15, resnet:he:CVPR16} and 3D recognition~\cite{pointnet}. Despite the rapid development of CNNs, forest methods 
based on random subspaces~\cite{randomsubspace:tin:1998} such as random forests~\cite{randomforest:breiman:ML01} and GBDTs~\cite{gbdt:friedman:2001} are still the dominant approaches for dealing with vectorized inputs in real-world applications.\footnote{\url{https://www.kaggle.com/amberthomas/kaggle-2017-survey-results}} Benefiting from ensemble predictors with methods such as bagging and boosting, forest methods are capable of making accurate and robust predictions.
However, training these models is computationally expensive, especially for large-scale datasets. 
Further, forest methods are mostly combinatorial rather than differentiable and they lack the capability of representation learning. On the other hand, CNNs integrate representation learning and classifier learning in an end-to-end fashion, with pervasive software (e.g., deep learning frameworks) and hardware (e.g., GPUs) support, which effectively work on large-scale datasets. Therefore, there have been deep forest methods which combine deep learning and forest learning to explore the advantages from both methods (e.g., \cite{mgbdt:feng:NIPS18}, \cite{neural-decision-forests:kontschieder:ICCV15}). However, the random subspace method in the context of deep learning has not been fully studied and also support from existing deep learning platforms has not been fully utilized in these previous works. Hence, one question arises: Can we enable random subspaces with end-to-end representation learning ability and better utilize deep learning software and hardware support?

Another interesting aspect is to examine the non-linearity in CNNs. 
By stacking layers of convolution and non-linearity, CNNs effectively learn discriminative representations. As one standard module in deep CNNs, global average pooling (GAP) summarizes linear statistics of the last convolution layer. Recently, many higher-order pooling (HOP) methods (e.g.,~\cite{bcnn:lin:ICCV15, cbp:gao:CVPR2016, isqrt:li:CVPR2018}) are proposed to learn higher-order, non-linear feature representations to replace GAP and have achieved impressive recognition accuracy. However, these HOP methods suffer from expensive computing costs because of the covariance calculation of very high dimensional matrices. Therefore, another question is: Can we add non-linearity to the linear GAP to achieve both good accuracy and high efficiency?

In this paper, we take a step towards addressing these two questions jointly. We propose a model called Neural Random Subspace (NRS), which is a deep learning based random subspace method. 
It realizes the random subspace method in the context of 
neural networks,
which well handles vectorized inputs (where CNNs do \emph{not} apply) and achieves both higher accuracy
(by combining ensemble and representation learning) and faster inference speed (by support from deep learning software and hardware) than conventional random subspace based forest methods, e.g., random subspaces and random forests. We show that such designs are attractive for many real-world tasks that deal with vector inputs.

Furthermore, NRS can be seamlessly installed after the GAP layer at the end of a CNN for image recognition, which non-linearly transforms the output of GAP. As a non-linear component to be encoded into CNNs, NRS is more efficient than HOP methods and achieves higher accuracy than standard GAP with negligible additional cost in terms of model parameters, FLOPs and inference time. Also, NRS can be installed across all layers in a CNN when integrated into Squeeze-and-Excitation (SE) modules~\cite{senet:hujie:arxiv} and it achieves comparable or better accuracy with fewer model parameters and FLOPs. Aside from 2D image recognition task, we also evaluate NRS on 3D classification tasks where it is used to non-linearly transform the output of the global feature encoders.

Experimental results validate the effectiveness of NRS. We achieve superior performance on 35 machine learning datasets when compared to previous forest methods. On document retrieval datasets, NRS achieves consistent improvements over various baseline algorithms. For 2D image recognition tasks, on the fine-grained benchmarks CUB-200-2011~\cite{cub200}, FGVC-aircraft~\cite{aircrafts} and Stanford Cars~\cite{cars}, by combining NRS we achieve $5.7\%$, $6.9\%$ and $7.8\%$ gains for VGG-16, respectively, with negligible increase in parameters, FLOPs and real running time. On ImageNet ILSVRC-12~\cite{ILSVRC2012:russakovsky:IJCV15}, integration of NRS into ResNet-18 achieves top-1/top-5 errors of $28.32\%/9.77\%$, which outperforms ResNet-18 by $1.92\%/1.15\%$ with negligible extra cost. For 3D recognition task on ModelNet40~\cite{ModelNet40}, NRS arises accuracy by $1.1\%$ for PointNet~\cite{pointnet} with minor extra complexities.

\section{Related Work}

\subsection{Forest Learning}
Forest learning is a powerful learning paradigm which often uses decision trees as its base learners. Bagging, boosting and the random subspace method (RSM), for instance, are the driving forces of random forests~\cite{randomforest:breiman:ML01}, GBDTs \cite{gbdt:friedman:2001} and random subspaces-based forests~\cite{randomsubspace:tin:1998}, respectively. In this section, we first review advances in the RSM and then deep forest learning methods.

\textbf{The random subspace method}: The RSM was first proposed in~\cite{randomsubspace:tin:1998}, which selects random subsets of features for base learners to construct decision forests. The RSM has several desirable attributes: (i) working in a reduced dimension space reduces the adverse consequences of the curse of dimensionality. (ii) RSM based ensemble classifiers could provide improved classification performance. (iii) it provides a stochastic and faster alternative to the optimal-feature-subset search algorithms. RSM approaches have achieved great success in various applications, including sparse representation~\cite{sparserep18:PR} and missing values~\cite{learn++10:PR}, etc.
However, the feasibility of implementing the RSM in the context of neural networks has not been fully explored, hence constitutes the focus of this paper. \cite{learn++10:PR} and \cite{first-episode:vyskovsky:FedCSIS16} use neural network ensembles based on random subspaces for schizophrenia classification and classification with missing data, respectively. Nevertheless, the RSM and neural networks are separate and end-to-end training for the RSM has not been fully studied in previous works. In contrast, our NRS unifies the RSM into neural networks in an end-to-end manner efficiently and effectively. Also, without the need to train multiple neural networks, our NRS integrates random subspace ensembles in a single model.

\textbf{Deep forest learning methods}: With the rapid development of deep learning, there have also been deep forest learning methods. 
There are mainly two reasons for combining deep learning and ensemble learning: 
(i) use deep learning hardware and software platforms to accelerate the training process, 
(ii) enrich ensemble methods with the capability of representation learning to boost the performance. 
On one hand, training those forest methods such as GBDTs is very time consuming, especially for large-scale and high dimensional datasets. To accelerate the training process, ThunderGBM~\cite{thundergbm:wen:19} proposes a GPU-based software to improve the efficiency of random forests and GBDTs. However, they are designed for specific algorithms and hardware, which is lack of generality in comparison with our NRS. On the other hand, it is widely recognized that the joint and unified way of learning feature representations together with their classifiers in deep CNNs greatly outperforms conventional feature descriptor \& classifier pipelines~\cite{resnet:he:CVPR16}. Hence, efforts have also been made to enrich forest methods with the capability of representation learning. mGBDTs~\cite{mgbdt:feng:NIPS18} learn hierarchical distributed representations by stacking several layers of regression GBDTs. 
However, these methods are not end-to-end trained and thus cannot be accelerated by the deep learning platforms. In contrast, our method integrates the RSM with end-to-end, data-driven representation learning capabilities with support from existing deep learning software and hardware platforms. NDF~\cite{neural-decision-forests:kontschieder:ICCV15} combines a single deep CNN with a random forest for image classification, where the outputs of the top CNN layer are considered as nodes of the decision tree and prediction loss is computed at each split node of the tree. 
Our work differs as follows: (\romannumeral1) We implement random subspaces rather than random forests in a novel and easy way. (\romannumeral2) We use existing CNN layers for the implementation and it makes our method easier to deploy to different platforms and devices. (\romannumeral3) Our method is more efficient by using group convolution for base leaner construction, which will be further studied in Sec 4.2.

\subsection{Non-linear representations in CNNs}
Statistics higher than first-order ones have been successfully used in both classic and deep learning based classification scenarios. For example, Fisher Vectors (FV)~\cite{fishervector:perronnin:ECCV10} use non-linear representations based on hand-crafted features. By replacing handcrafted features with outputs extracted from CNNs pre-trained on ImageNet~\cite{ILSVRC2012:russakovsky:IJCV15}, these models achieve state-of-the-art results on many recognition tasks~\cite{filterbank:compoi:CVPR15}. In these designs, representation and classifier training are not jointly optimized and end-to-end training has not been fully studied. \cite{bcnn:lin:ICCV15} proposes a bilinear CNN (B-CNN) that aggregates the outer products of convolutional features from two networks and allows end-to-end training for fine-grained visual classification. 
\cite{isqrt:li:CVPR2018} proposes an iterative matrix square root normalization (iSQRT) method for fast training of global covariance pooling networks. These works have shown that higher-order, non-linear feature representations based on convolution outcomes achieve impressive improvements over the classic linear representations. However, they suffer from the expensive computational overhead because these methods depend heavily on spectral or singular value decomposition of very high dimensional covariance matrices. 
Contrary to previous higher-order methods, our NRS learns non-linear feature representations with only negligible increase in parameters, FLOPs and real running time while achieving competitive performance.


\section{Neural Random Subspace}
\label{sec:NRS}
In this section, we propose the NRS module, which mainly consists of random permutations and group convolutions. 
We show that it essentially resembles an ensemble of one-level decision trees where each tree learns from a subset of the features, i.e., a random subspace and we name it Neural Random Subspace (NRS). First we revisit the RSM and then introduce our method.

\subsection{RSM recap}
Let $D=\{(\boldsymbol{x}^{(i)}, y^{(i)})|i=1,\cdots,n\}$ denote a dataset, where each $\boldsymbol{x}^{(i)} \in \mathbb{R}^d$ represents an input instance with its associated label $y^{(i)}$, $n$ is the number of instances and $d$ is the number of features. Each instance $\boldsymbol{x}^{(i)}$ is represented by a $d$-dimensional feature vector $(x^{(i)}_1,\cdots,x^{(i)}_d)$. The RSM is organized in the following way:

\noindent 1. Repeat for $j=1,\cdots, T$:

(a) randomly selects $r<d$ features from the $d$-dimensional dataset $D$ and thus obtains the $r$-dimensional random subspace $\tilde{D}^j$ of the original feature space.

(b) constructs classifiers $C^j(\boldsymbol{x})$ in the random subspace $\tilde{D}^j$.

\noindent 2. combine the results of the $T$ classifiers $C^j(\boldsymbol{x})$ $(j=1,\cdots,T)$. 

As can be seen, there are two key points to implement the RSM in the context of neural networks: how to construct different random subspaces and how to construct corresponding base learners? In the next subsection, we introduce the two main techniques in our NRS called \emph{expansion} and \emph{aggregation}, which answer these two questions correspondingly. 

\subsection{Module architecture}
\label{sec:3.1}

Our goal is to build an ensemble of neural classifiers based on random subspace, which combines the advantages of ensemble learning and deep learning. To achieve this, we propose a novel architecture named NRS (in Figure~\ref{fig:network}). Our NRS mainly contains two parts: 

\begin{figure}[t]
	\centering
	\includegraphics[width=\columnwidth]{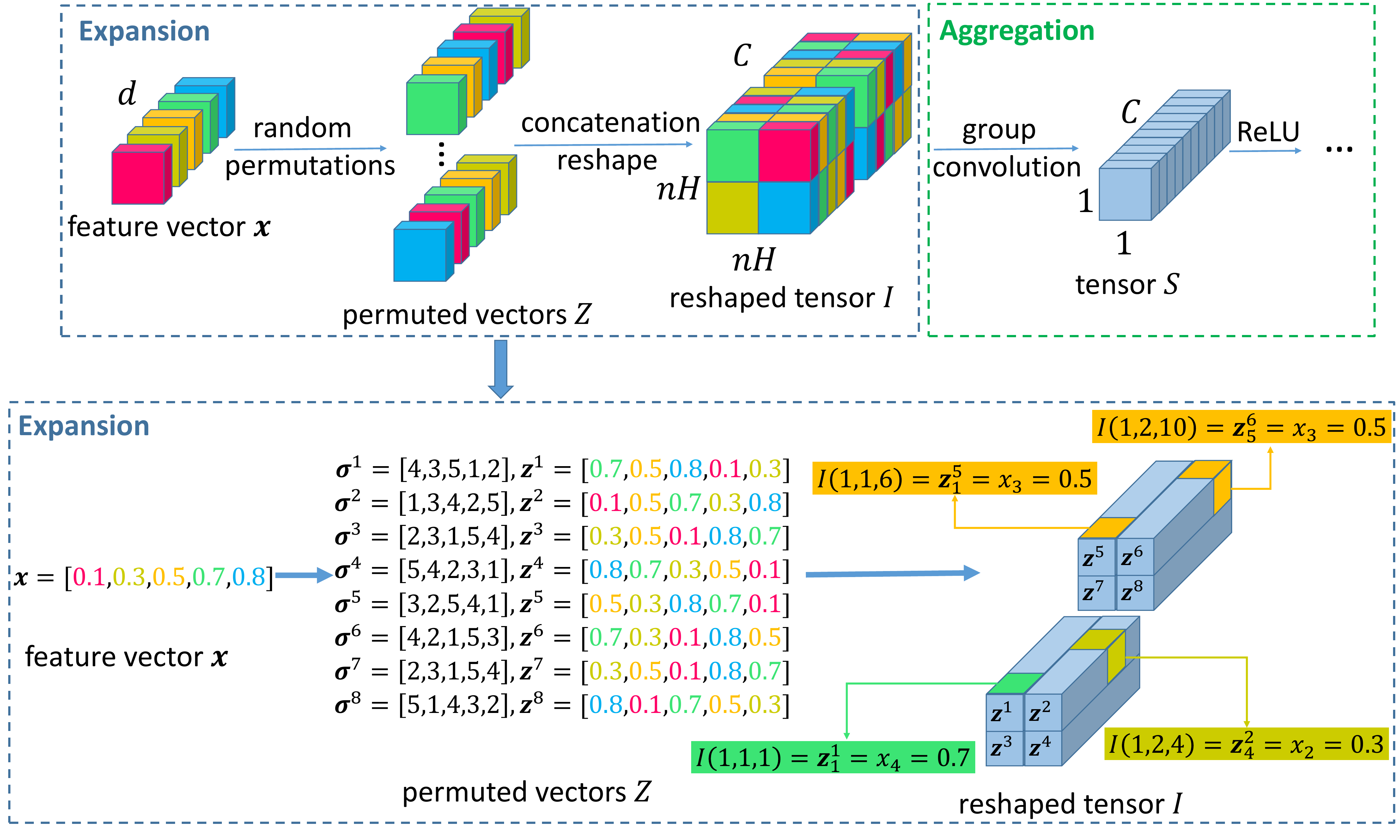}
	\caption{NRS architecture. In this figure we have input feature vector $\boldsymbol{x}=(x_1,\cdots,x_5)$ where different features are marked different colors, e.g., the red color corresponds to $x_1$, etc. Here we set the expansion rate in depth $nMul=2$, the spatial width and height $nH=2$. Hence, we have $C=nMul\times{d}=10$ and $M=nH\times{nH}\times{nMul}=8$ permuted vectors. We show a concrete example in the lower part of this figure for better illustration of the permutation and reshaping process. (This figure is best viewed in color.)} 
	\label{fig:network}
\end{figure}

(i) \textbf{Expansion:} Firstly, NRS takes as input a $d$-dimensional feature vector $\boldsymbol{x}\in\mathbb{R}^d$ and produces a $nH\times{nH}\times{C}$ tensor $I$ where $nH$ is the spatial width and height of a square output tensor, $C={nMul}\times{d}$ is the number of output channel (output depth) and $nMul$ is the expansion rate in depth. We now introduce this process in detail. For the $d$-dimensional input feature vector $\boldsymbol{x}$, we first generate $M$ random permutations of it 
by reordering its elements where $M=nH\times{nH}\times{nMul}$, as depicted by the first arrow in Figure~\ref{fig:network}. We represent the $M$ permuted element orders as $\boldsymbol{\sigma}^1,\cdots, \boldsymbol{\sigma}^M$ (e.g., a reverse permutation order $\boldsymbol{\sigma}=[d,\cdots,2,1]$). This results in a set of randomly permuted vectors $Z=\{z^t\}_{t=1,\cdots,M}$ from $\boldsymbol{x}$ correspondingly, where $z^t$ is generated by $t$-th permutation order $\boldsymbol{\sigma}^t$:
\begin{equation}
\label{eq:zt}
 \boldsymbol{z}^t = ({x}_{\boldsymbol{\sigma}^t_1}, \cdots, {x}_{\boldsymbol{\sigma}^t_d}), \quad t=1,\cdots,M \,,
\end{equation}
where the $i$-th entry of $t$-th permutation is ${x}_{\boldsymbol{\sigma}^t_i}$. Then, we concatenate these $M$ permuted feature vectors and reshape it into a 3D tensor $I\in\mathbb{R}^{nH\times{nH}\times{C}}$. We denote the entry at $i$-th row, $j$-th column and $k$-th channel in $I$ as $I(i,j,k)$ and it essentially corresponds to the $s$-th element in $\boldsymbol{z}^t$:
\begin{equation}
\label{I(i,j,k)}
I(i,j,k) = \boldsymbol{z}^t_s={x}_{\boldsymbol{\sigma}^t_{s}}\,,
\end{equation}
where
\begin{equation}
\begin{split}
\label{t-permutation}
s &= k \bmod d \,,\\
t = \lfloor{k/d}\rfloor\times{nH}&\times{nW}+(i-1)\times{nW}+j \,.
\end{split}
\end{equation}
In Figure~\ref{fig:network} we show the scenario when $nH=2$ and $nMul=2$.

(ii) \textbf{Aggregation:} We subsequently process the tensor $I$ with a group convolution layer of kernel size $(nH, nH)$, output channel numbers $C$ and group numbers $\lfloor{c/nPer}\rfloor$ with no padding. Here $nPer$ denotes the number of channels per group in the group convolution layer. This results in a new 1D tensor $S\in\mathbb{R}^{1\times 1\times c}$ followed by ReLU non-linearity, as shown in Figure~\ref{fig:network}. Finally, we feed it into fully connected (FC) layers plus a softmax layer for classification tasks. Without further modifications on the model, our NRS is a generalised module that are compatible with most network models, we have also shown that in the experiments sections. Additionally we can combine multiple group convolution layers with smaller kernel sizes to make NRS deeper.


\subsection{Neural random subspace via CNN implementation}

A $d$-dimensional input vector $\boldsymbol{x}$, can be either a handcraft feature in traditional machine learning or pattern recognition tasks or a learned representation extracted by CNNs (e.g., the output of a GAP layer). In NRS, we first transform such vector into a 3D tensor $I$ by random permutation and concatenation.

Tensor $I$ composes of a set of 2D feature maps $I=\{I^k\}_{k=1,\cdots, C}$. $I^k$, of size $nH\times{nH}$, is the feature map of the $k$-th channel. Each feature map $I^k$ consists of $nH\times{nH}$ features that are randomly selected from the original feature vector $\boldsymbol{x}$. In other words, each feature map $I^k$ corresponds to a random subset of features, that is, a random subspace. Moreover, each group convolution filter which randomly chooses $nH\times{nH}\times{nPer}$ features 
and the consequent ReLU layer which acts upon an attribute (i.e., a linear combination of these randomly selected features) 
can be considered as a one-level oblique decision tree~\cite{oblique:murthy:IJCAI93}. 
We validate this statement as following. 

For simplicity, we show the situation when the group convolution layer has $nPer=1$ (i.e., a depthwise convolution) in Figure~\ref{fig:forest}. We use $W^k$ to denote the weights of $k$-th filter, $S^k$ to denote the $k$-th channel of $S$. 

Based on the notations from Equation~\eqref{I(i,j,k)}, the $k$-th channel of $S$ is convoluted on the $k$-th feature map of $I$ with the kernel $W^k$:
\begin{figure}[t]
	\centering
	\includegraphics[width=0.9\columnwidth]{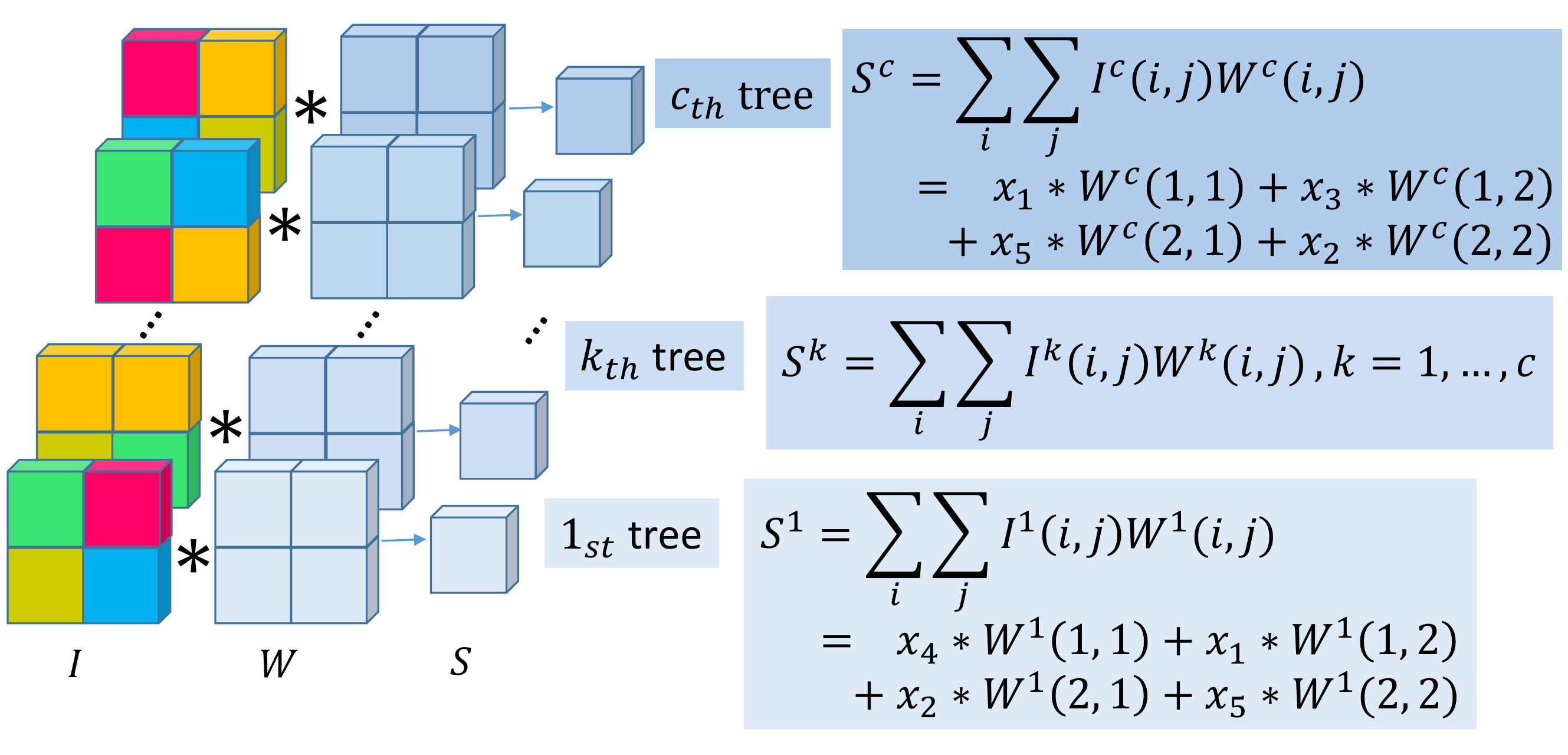}
	\caption{Group convolution makes an ensemble of one-level trees. Each square in different color corresponds to different feature in input feature vector $\boldsymbol{x}$ of 5 dimensions in Figure~\ref{fig:network}.}
	\label{fig:forest}
\end{figure}


\begin{equation}
\label{Sk}
S^k = \sum_{i}\sum_{j}I^k(i,j)W^k(i,j) =\sum_{i}\sum_{j}{x}_{\boldsymbol{\sigma}_{s}^t}\times{W^k(i,j)}\,.
\end{equation}
Let $f(\cdot)$ denote the ReLU function, the output of NRS is computed as:
\begin{equation}
\label{Tk}
f(S^k) = \left\{
\begin{array}{rcl}
S^k&  & {S^k \geq 0}\\
0\phantom{0} && {S^k < 0}
\end{array} \right.
\end{equation}
From Equation~\eqref{Tk} and Figure~\ref{fig:forest} we can see that each convolution filter $W^k$ along with the subsequent ReLU resembles a one-level tree which outputs a linear combination of randomly selected features and then a decision based on it. Hence, all convolution filters form an ensemble with $C$ different one-level trees. Actually, if we use $1\times1$ depthwise group convolution where $nH=1$, each base learner reduces to a decision stump which learns with a single feature. In conclusion, the random permutation operation acts as resampling and group convolution is used for aggregation. These operations in effect construct random spaces in the context of neural networks where each base learner learns from one random subspace. Finally, outputs from those random subspaces are combined for final classification through a combination function realised by FC layers.

It is worth noting that the previous decision forests based on random subspaces use bootstrapping to generate feature subsets, in which there is a chance that not all the features will be used~\cite{bagging:breiman:ML96}. Instead, in NRS every feature occurs equal times ($M$ times) therefore it is naturally guaranteed that each feature will be considered in the final ensemble. Meanwhile, the injected randomness in NRS guarantees the difference of each random subspace.

Moreover, when we increase $nMul$, the number of channels $C$ gets larger and we will get more group convolution filters. Hence, from Equation~\eqref{Sk}, more random subspaces are integrated into the ensemble correspondingly. 
We can further rise $nPer$ and $nH$ to increase the amount of features in each random subspace, i.e., the capacity of each base learner. 
Finally, by stacking more group convolution layers, 
we can make NRS deeper. In Sec~\ref {sec:ablation-study} we conduct studies about $nMul, nPer$, $nH$ and the number of group convolution layers in NRS.

\subsection{Computational Complexity}
In the following, we show that NRS is efficiency friendly in terms of the parameters and computational complexity. A standard convolution layer process an input $\mathbf{F}\in\mathbb{R}^{H_{in}\times{H_{in}}\times{C_{in}}}$ 
into an output $\mathbf{G}\in\mathbb{R}^{H_{out}\times{H_{out}}\times{C_{out}}}$, where $C_{in}$ and $C_{out}$ are the number of input and output channels respectively, $H_{in}$ and $H_{out}$ are the spatial width and height of a square input and output tensor, respectively. Notice that in the standard convolution layer the output tensor has the same spatial dimensions as the input, i.e., $H_{in}=H_{out}$. The layer is parameterized by convolution kernel $\mathbf{K}$ of size $(K, K, C_{in}, C_{out})$ where $K$ is the spatial dimension of the kernel. Hence the computational cost of this standard convolution is:
\begin{equation}
\label{eq:conv_cost}
    K^2\times H_{in}^2\times{C_{in}}\times{C_{out}}.    
\end{equation}

Group convolution reduces this cost by dividing the $C_{in}$ input channels into $G$ non-overlapping groups. After applying filters over each group, group convolution generates $C_{out}$ output channels by concatenating the outputs of each group. Each group is parameterized by convolution kernel(s) of size $(K, {K}, {\frac{C_{in}}{G}}, {\frac{C_{out}}{G}})$ and there are $G$ groups in total. Hence, group convolution has 
a number of total parameters of:
\begin{equation}
\label{eq:group-parameters}
\frac{K^2\times{C_{in}}\times{C_{out}}}{G}, 
\end{equation}
which is reduced by $G$ times when compared with standard convolution. The computational cost of the group convolution is:
\begin{equation}
\label{eq:group-cost}
    \frac{K^2\times{H_{in}^2}\times{C_{in}}\times{C_{out}}}{G},
\end{equation}
which is also decreased by a factor of $1/G$. When $G=C_{in}$, the maximum reduction is achieved and group convolution is called depthwise convolution.  

In our NRS, we adopt depthwise convolution in most of our experiments (i.e., $nPer=\frac{C_{in}}{G}=1$) and we set $C_{in}=C_{out}=C$, where $nPer$ and $C$ are defined in Sec~\ref{sec:3.1}. Also, we set the spatial dimension of the kernel equals the input ($K=H_{in}$) without padding, hence the output spatial dimension $H_{out}=1$. Our NRS has a total number of parameters of:
\begin{equation}
\label{eq:nrs-parameters}
K^2\times{C_{out}}
\end{equation}
which is further reduced by $C_{in}$ times when compared with standard convolution. Also, NRS has the computation cost of:
\begin{equation}
\label{eq:nrs-cost}
    K^2\times{C_{out}}
\end{equation}
which is reduced by $C_{in}\times{H_{in}}\times{H_{in}}$ times when compared with standard convolution and ${H_{in}}\times{H_{in}}$ times when compared with regular depthwise convolution. Hence, NRS is efficient in terms of both parameters and computational complexity.

As can be seen, the group convolution design NRS has at least two benefits:
\begin{enumerate}[(i)]
    \item The non-overlapping idea in group convolution is naturally consistent with the random subspace method. In RSM, each base learner learns from a random subset of features and there is no interaction between base learners. However, in standard convolutions, all features will be utilized by each convolution filter. In contrast, in group convolutions, each group convolution filter only utilizes features in its own group and there is no overlapping between groups, which is consistent with the random subspace idea.
    \item Group convolution is efficient in terms of both parameters and computational complexity. We will show that NRS actually adds negligible parameters and FLOPs through experimental results in the next section.
\end{enumerate}

\section{Experimental Results}
In the following section, we will empirically evaluate the effectiveness of our NRS module. On one hand, for vectorized inputs, we compare our method with other competitive forest methods on 34 machine learning classification datasets as well as 1 multivariate regression dataset SARCOS in Sec~\ref{sec:machine learning datasets}. Moreover, we also evaluate our NRS on the challenging document retrieval task Microsoft 10K and Microsoft 30K~\cite{mslrweb} in Sec~\ref{sec:retrieval}. On the other hand, NRS can be integrated into CNNs for improving non-linear capability either at the end of or across all layers in the network. We conduct experiments on CIFAR-10~\cite{cifar}, CIFAR-100~\cite{cifar}, fine-grained visual categorization benchmarks and the large-scale ImageNet ILSVRC-12~\cite{ILSVRC2012:russakovsky:IJCV15} task with five widely used deep models: MobileNetV2~\cite{mobilenetv2:sabdker:CVPR18}, VGG~\cite{vgg:simonyan:ICLR15}, ResNet~\cite{resnet:he:CVPR16}, Inception-v3~\cite{inceptionv3:szegedy:cvpr16} and SENet~\cite{senet:hujie:arxiv}. Aside from 2D image recognition tasks, we also evaluate our NRS on the 3D recognition task on ModelNet40~\cite{ModelNet40} in Sec~\ref{sec:3d} with three widely used baselines: PointNet\cite{pointnet} , PoinetNet++\cite{pointnet++} and DGCNN~\cite{dgcnn}. All our experiments were conducted using PyTorch on Tesla M40 GPUs and we have made our code publicly available\footnote{\url{https://github.com/CupidJay/NRS_pytorch}}. 

\subsection{Overview}

For machine learning classification and regression datasets, a brief description of them including the train-test split, the number of categories and feature dimensions is given in Table~\ref{tab:ml-dataset-extensive}. For document retrieval datasets, we use the popular benchmark Microsoft 10K and Microsoft 30K~\cite{mslrweb}. Microsoft 10K consists of 10,000 queries and 1,200K documents while Microsoft 30K has 31,531 queries and 3,771K documents.

For image datasets, CIFAR-10~\cite{cifar} consists of 50,000 training images and 10,000 test images in 10 classes and CIFAR-100~\cite{cifar} is just like the CIFAR-10, except it has 100 classes containing 600 images for each class. For fine-grained categorization, we use three popular fine-grained benchmarks, i.e.,  CUB-200-2011 (Birds)~\cite{cub200}, FGVC-aircraft (Aircraft)~\cite{aircrafts} and Stanford Cars (Cars)~\cite{cars}. The Birds dataset contains 11,788 images from 200 species, with large intra-class but small inter-class variations. The Aircraft dataset includes 100 aircraft classes and a total of 10,000 images with small background noise but higher inter-class similarity. The Cars dataset consists of 16,185 images from 196 categories. For large-scale image classification, we adopt the ImageNet ILSVRC-12 dataset~\cite{ILSVRC2012:russakovsky:IJCV15} with 1,000 object categories. The dataset contains 1.28M images for training, 50K images for validation and 100K images for testing. As in \cite{resnet:he:CVPR16}, we report the results on the validation set.

For the 3D object recognition task, we use ModelNet40~\cite{ModelNet40} as the benchmark dataset. It contains 40 classes of synthesized CAD models, where 9,823 objects are used for training and 2,464 objects for testing. We randomly sample 1,024 points from the mesh faces as the point cloud representation for each object. 

\subsection{Machine learning datasets} \label{sec:machine learning datasets}
We compare NRS with forest methods, e.g., decision forests based on random subspaces (RSs)~\cite{randomsubspace:tin:1998}, random forests (RFs)~\cite{randomforest:breiman:ML01} and GBDTs~\cite{gbdt:friedman:2001} in terms of accuracy, training/testing time and model size. Furthermore, because we use NRS with 2 FC layers on these machine learning datasets, we also compare it with multi-layer perceptrons (MLP). Our NRS has one more convolution layer than MLP with 2 FC layers (denoted as MLP-2) and hence for fair comparisons we compare with both MLP with 2 FC layers and MLP with 3 FC layers (denoted as MLP-3). Notice that when using dropout~\cite{dropout:Srivastava:ICML14} at the input layer in MLP (denoted as MLP-D), it can be considered as an ensemble of neural networks trained from different subsets of features and we also compare with it.

\begin{table}[!htbp]
	\caption{Statistics of the machine learning datasets reported in the paper. The above 34 datasets are classfication datasets and SARCOS is a regression dataset.}
	\label{tab:ml-dataset-extensive}
	\centering
	\small
	\setlength{\tabcolsep}{2.8pt}
	\renewcommand{\arraystretch}{0.5}
	\begin{tabular}{l|r|r|r|r|r|c|c}
		\hline
		\multirow{2}{*}{Datasets} &\multicolumn{4}{c|}{Statistics}  & \multicolumn{3}{c}{NRS setting}\\
		\cline{2-8}
		     & \# Category   & \# Training & \# Testing & \# Dim & $nMul$ & $nPer$ &$nH$\\
		
		\hline
		satimage  & \phantom{0}6  & \phantom{0}4435& \phantom{0}2000 & \phantom{0}\phantom{0}36 &20&1&3\\
		GISETTE &\phantom{0}2& \phantom{0}6000 & \phantom{0}1000  &5000&10&1&3 \\
		MNIST  & 10   & 60000 & 10000  & \phantom{0}780&16&1&3\\
		letter  & 26 & 15000 & \phantom{0}5000 & \phantom{0}\phantom{0}16&100&1&3\\
		USPS  & 10 & \phantom{0}7291 & \phantom{0}2007 & \phantom{0}256&30&1&3\\
		yeast  & 14 & \phantom{0}1500 & \phantom{0}\phantom{0}917 & \phantom{0}\phantom{0}\phantom{0}8&20&1&3\\
		dna & 3 & 1400 & 1186 & 180&5&1&3\\
		ijcnn1 & 2 & 35000 & 91701 & 22&10&1&3\\
		pendigits & 10 & 7494 & 3498 & 16&20&1&3\\
		poker & 10 & 25010 & 1000000 & 10&50&1&3\\
		protein & 3 & 14895 & 6621 & 357&2&1&3\\
		segment& 7 & 1617&693&19&30&1&3\\
		SVHN&10&73257&26032&3072&1&1&3\\
		CIFAR-10&10&50000&10000&3072&1&1&3\\
		connect-4&3&47289&20268&126&5&1&3\\
		SensIT&3&78823&19075&50&20&1&3\\
		splice&2&1000&2175&60&10&1&3\\
		a1a&2&1605&30956&123&10&1&3\\
		a9a&2&32561&16281&123&10&1&3\\
		aloi&1000&75600&32400&128&10&1&3\\
		cod-rna&2&59535&271617&8&50&1&3\\
		covtype&2&406708&174304&54&20&1&3\\
		SUSY&2&3500000&1500000&18&20&1&3\\
		australian&2&483&207&14&20&1&3\\
		breast-cancer&2&478&205&10&50&1&3\\
		fourclass&2&603&259&2&50&1&3\\
		german&2&700&300&24&30&1&3\\
		diabetes&2&537&231&8&50&1&3\\
		heart&2&189&81&13&30&1&3\\
		vehicle&4&592&254&18&30&1&3\\
		sonar&2&145&63&60&10&1&3\\
		glass&6&149&65&9&50&1&3\\
		ionosphere&2&245&106&34&20&1&3\\
		phishing&2&7738&3317&68&10&1&3\\
		\hline
		SARCOS & - & 44484 & \phantom{0}4449 & \phantom{0}\phantom{0}21&40&1&3\\
		\hline
	\end{tabular}
\end{table}

\textbf{Implementation details:} We build NRS by 1 group convolution layer and 2 FC layers with batch normalization (BN) in all datasets. We construct MLP-2 and MLP-3 with BN in the same way. For MLP-D, we add dropout at the input layer with $p=0.8$ as is done in~\cite{dropout:Srivastava:ICML14} and other settings remain the same as MLP-2 and MLP-3. We split $10\%$ of the training data for validation to determine the total epochs separately for each dataset. We train all networks for 20$\sim$50 epochs, using Adam~\cite{adam} as optimizer and initializing learning rate to 1e-4. In Table~\ref{tab:ml-dataset-extensive}, we set $nPer$ to 1 and $nH$ to 3 for all these datasets for simplicity. Considering feature dimensionalities among different datasets, we set different $nMul$ for these datasets to ensure that the product of dimensionalities and $nMul$ is within a relative reasonable interval to save computing resources, as shown in Table~\ref{tab:ml-dataset-extensive}. For MLP-2, MLP-3, MLP-D, RSs, RFs and GBDTs, we carefully tune the parameters through 5-fold cross-validation on the training set and choose the best parameters for them in each dataset. We report the mean accuracy and standard deviation of 5 trials for all datasets except yeast, which is evaluated by 10-fold cross-validation. 

\begin{figure}[t]
	\centering
	\includegraphics[width=\columnwidth]{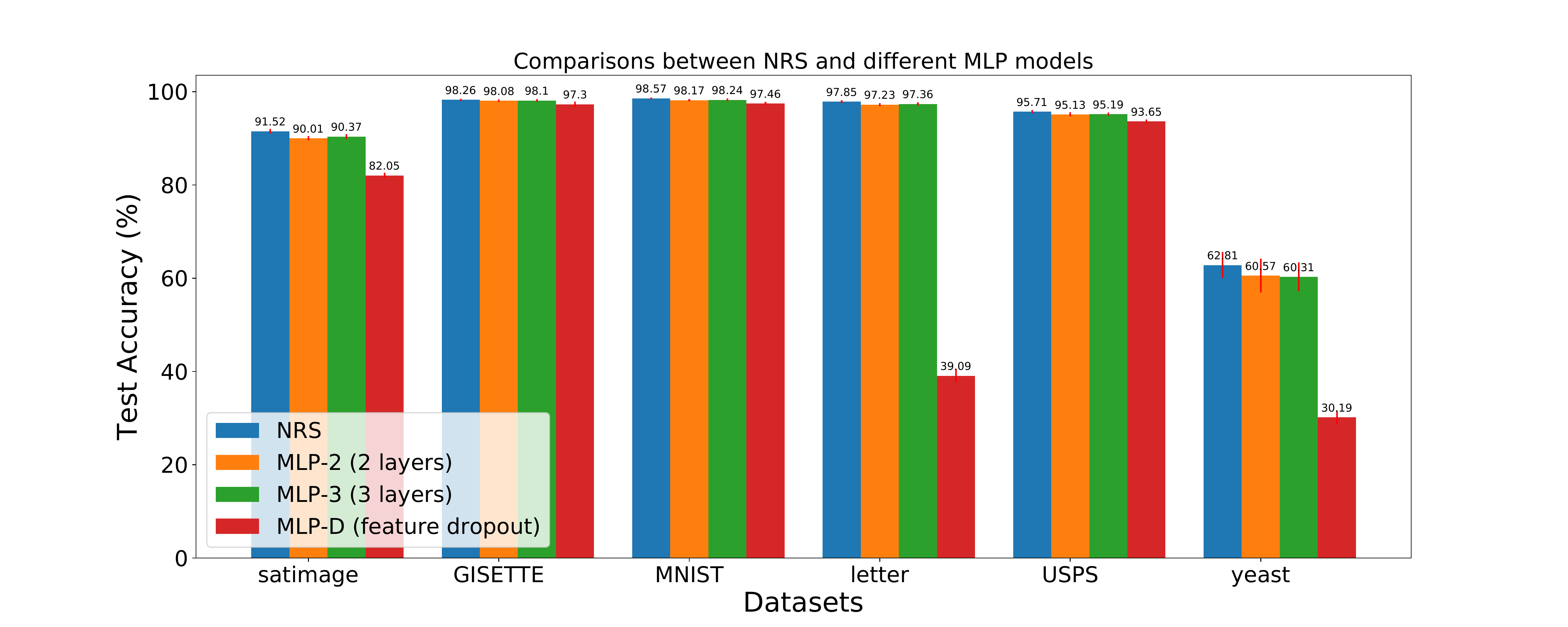}
	\caption{Comparison between NRS and several MLP variants on satimage, GISETTE, MNIST, letter, USPS and yeast. We plot the average accuracy and standard deviation of 5 trails at each bar.}
	\label{fig:MLP_variants}
\end{figure}

We choose the first 6 datasets satimage, GISETTE, MNIST, letter, USPS and yeast to compare the performance of NRS and several MLP variants, namely MLP-2, MLP-3 and MLP-D, as shown in Figure~\ref{fig:MLP_variants}. For model size, speed and accuracy comparisons, we choose the \emph{largest} dataset SUSY, the \emph{highest} dimensional dataset GISETTE and 2 medium-sized dataset MNIST and letter and we use some different experimental settings for algorithms in Table~\ref{tab:ml-dataset-speed-compare}. It is hard to make an absolutely fair comparison and for better trade-off between model size, speed and accuracy, we reduce the number of trees for random subspaces (RSs), random forests (RFs), GBDTs and NDF and the parameter $nMul$ for NRS correspondingly. In Table~\ref{tab:ml-dataset-speed-compare}, we use the same settings as in Table~\ref{tab:ml-dataset-res} except that we set $nMul$ to 1, 5, 50 and 10 for NRS for GISETTE, MNIST, letter and SUSY, respectively, for better model size, speed and accuracy trade-off.  Correspondingly, we reduce the number of trees from 500 to 100 for RSs, RFs, GBDTs and from 100 to 50 for NDF for faster speed and smaller size. We also use 5-fold cross validation on the train set and choose the best value for other parameters. We record total training time on the train set and inference time on the test set in seconds.

\begin{table}[!htbp]\small
	\caption{Accuracy(\%) on machine learning benchmarks. We report the average accuracy and standard deviation of 5 trails. NRS and MLP are the results of last epoch. $\bullet/\circ$ indicates that our NRS is significantly better/worse than the corresponding method (pairwise t-tests at 95\% significance level). `N/A' means that no results were obtained after running out 250000 seconds (about 3 days). The last row is the results on the regression dataset SARCOS and * denotes that \cite{ant:tanno:ICML19} didn't report the standard deviation.}
	\label{tab:ml-dataset-res}
	\centering
	\scriptsize
	\setlength{\tabcolsep}{2.2pt}
	\begin{tabular}{l|c|c|c|c|c|c}
		\hline
		Datasets     & NRS (ours) & MLP & NDF~\cite{neural-decision-forests:kontschieder:ICCV15} & RSs & RFs & GBDTs \\
		\hline
		satimage  & \textbf{91.52$\pm$0.31} &90.01$\pm$0.31$\bullet$ &89.71$\pm$0.31$\bullet$ & 90.97$\pm$0.08$\bullet$  & 91.01$\pm$0.35$\bullet$ & 89.26$\pm$0.04$\bullet$    \\
		GISETTE & \textbf{98.26$\pm$0.05}&98.08$\pm$0.12$\bullet$&97.24$\pm$0.29$\bullet$ & 95.72$\pm$0.12$\bullet$  & 96.98$\pm$0.13$\bullet$ & 97.18$\pm$0.04$\bullet$   \\
		MNIST   & \textbf{98.57$\pm$0.03} &98.17$\pm$0.07$\bullet$&98.36$\pm$0.12$\bullet$ & 96.83$\pm$0.03$\bullet$  & 96.96$\pm$0.08$\bullet$ & 96.56$\pm$0.07$\bullet$  \\
		letter  & \textbf{97.85$\pm$0.10}&97.23$\pm$0.17$\bullet$&97.08$\pm$0.17$\bullet$ & 96.94$\pm$0.11$\bullet$ & 96.14$\pm$0.10$\bullet$ & 94.66$\pm$0.01$\bullet$\\
		USPS  & \textbf{95.71$\pm$0.17} &95.13$\pm$0.26$\bullet$ & 94.99$\pm$0.24$\bullet$ & 92.81$\pm$0.04$\bullet$& 93.80$\pm$0.19$\bullet$ & 92.83$\pm$0.03$\bullet$ \\
		yeast  & \textbf{62.81$\pm$2.61} & 60.57$\pm$3.45\phantom{0} &60.31$\pm$3.37$\bullet$ &58.40$\pm$2.90\phantom{0} &62.81$\pm$3.47\phantom{0} & 60.71$\pm$2.35\phantom{0} \\
		dna  & 94.91$\pm$0.22 &92.56$\pm$0.44$\bullet$ & 93.12$\pm$0.20$\bullet$ & 94.60$\pm$0.11$\bullet$ & 93.64$\pm$0.27$\bullet$   & \textbf{95.53$\pm$0.00}$\circ$\\
		ijcnn1  & 98.34$\pm$0.11 &\textbf{98.55$\pm$0.16}$\circ$&98.51$\pm$0.19\phantom{0} & 97.15$\pm$0.06$\bullet$ & 96.76$\pm$0.09$\bullet$   & 96.17$\pm$0.05$\bullet$\\
		pendigits  & \textbf{98.03$\pm$0.19} &97.11$\pm$0.18$\bullet$ &97.55$\pm$0.12$\bullet$ & 96.79$\pm$0.11$\bullet$ & 96.46$\pm$0.06$\bullet$   & 96.13$\pm$0.01$\bullet$\\
		poker  & 79.28$\pm$1.29 &70.13$\pm$1.60$\bullet$ &68.43$\pm$0.41$\bullet$ & 74.29$\pm$0.30$\bullet$ & 64.97$\pm$0.26$\bullet$   & \textbf{88.13$\pm$0.23}$\circ$\\
		protein  & \textbf{69.88$\pm$0.31} &67.65$\pm$0.19$\bullet$ &69.47$\pm$0.17\phantom{0} & 68.42$\pm$0.20$\bullet$ & 68.75$\pm$0.25$\bullet$   & 68.93$\pm$0.01$\bullet$\\
		segment  & 97.26$\pm$0.27 &96.45$\pm$0.60$\bullet$ &95.04$\pm$0.35$\bullet$ & \textbf{97.34$\pm$0.20}\phantom{0} & 94.29$\pm$0.20$\bullet$   & 96.97$\pm$0.01$\bullet$\\
		SVHN  & \textbf{82.16$\pm$0.29} &78.07$\pm$3.12$\bullet$ &78.96$\pm$0.54$\bullet$ & 68.06$\pm$0.17$\bullet$  & 70.33$\pm$0.13$\bullet$   & 71.74$\pm$0.20$\bullet$\\
		CIAFR-10  & \textbf{56.11$\pm$0.21} &46.42$\pm$2.28$\bullet$ &54.04$\pm$0.41$\bullet$ & 47.06$\pm$0.35$\bullet$ & 48.99$\pm$0.07$\bullet$   & 54.12$\pm$0.01$\bullet$ \\
		connect-4  & 85.70$\pm$0.15 &84.95$\pm$0.20$\bullet$ &\textbf{86.32$\pm$0.11}$\circ$ & 83.52$\pm$0.09$\bullet$ & 82.81$\pm$0.11$\bullet$   & 80.34$\pm$0.01$\bullet$\\
		SensIT  & \textbf{80.39$\pm$0.22} &80.03$\pm$0.63\phantom{0} &70.30$\pm$0.73$\bullet$ & 80.13$\pm$0.03$\bullet$ & 79.89$\pm$0.06$\bullet$   & 80.12$\pm$0.01$\bullet$\\
		splice  & 93.25$\pm$0.50 &88.43$\pm$0.81$\bullet$ &91.16$\pm$0.22$\bullet$ & \textbf{97.03$\pm$0.14}$\circ$ & 96.68$\pm$0.15$\circ$   & 96.78$\pm$0.01$\circ$\\
		a1a  & \textbf{84.33$\pm$0.08} &81.86$\pm$0.22$\bullet$ &83.18$\pm$0.26$\bullet$ & 82.13$\pm$0.06$\bullet$ & 83.06$\pm$0.10$\bullet$   & 83.61$\pm$0.00$\bullet$\\
		a9a  & 85.06$\pm$0.04 &82.54$\pm$0.13$\bullet$ &84.84$\pm$0.05$\bullet$ & 83.52$\pm$0.03$\bullet$ & 84.77$\pm$0.05$\bullet$   & \textbf{85.36$\pm$0.03}$\circ$\\
		aloi  & 95.76$\pm$0.09 &95.10$\pm$0.08$\bullet$ &N/A & 95.61$\pm$0.05$\bullet$ & \textbf{95.86$\pm$0.03}$\circ$   & N/A\\
		cod-rna  & 96.71$\pm$0.04 &96.69$\pm$0.05\phantom{0} &96.48$\pm$0.04$\bullet$ & 95.94$\pm$0.06$\bullet$ & 96.65$\pm$0.01$\bullet$   & \textbf{96.85$\pm$0.01}$\circ$\\
		covtype  & 96.08$\pm$0.06 &94.88$\pm$0.13$\bullet$ &93.62$\pm$0.15$\bullet$ & \textbf{97.30$\pm$0.03}$\circ$ & 95.98$\pm$0.02$\bullet$   & 95.73$\pm$0.01$\bullet$\\
		SUSY  & \textbf{80.47$\pm$0.01} &80.44$\pm$0.01$\bullet$ & 79.92$\pm$0.01$\bullet$ & 79.89$\pm$0.02$\bullet$ & 80.16$\pm$0.01$\bullet$   & 80.35$\pm$0.00$\bullet$\\
		australian  & 87.44$\pm$0.75 &86.09$\pm$0.89$\bullet$ & 86.67$\pm$1.13\phantom{0} & 86.09$\pm$0.56$\bullet$ & 86.86$\pm$0.36\phantom{0}   & \textbf{87.92$\pm$0.00}\phantom{0}\\
		breast-cancer  & \textbf{97.46$\pm$0.57} & 96.88$\pm$0.79\phantom{0} &96.10$\pm$0.31$\bullet$ & 95.51$\pm$0.20$\bullet$ & 96.20$\pm$0.20$\bullet$   & 95.61$\pm$0.00$\bullet$\\
		fourclass  &99.54$\pm$0.45 & 99.08$\pm$0.67\phantom{0} &\textbf{99.61$\pm$0.10}\phantom{0} & 97.14$\pm$1.70$\bullet$ & 95.67$\pm$0.15$\bullet$   & 98.46$\pm$0.00$\bullet$\\
		german  & 76.60$\pm$0.53 &75.20$\pm$0.72$\bullet$ & 74.00$\pm$0.47$\bullet$ & 75.33$\pm$0.47$\bullet$ & 71.40$\pm$0.39$\bullet$   & \textbf{77.33$\pm$0.00}$\circ$\\
		diabetes  & 74.89$\pm$0.61 & 74.11$\pm$1.76\phantom{0} &75.32$\pm$1.02\phantom{0} & 71.86$\pm$0.61$\bullet$ & \textbf{75.84$\pm$0.32}$\circ$   & 71.86$\pm$0.00$\bullet$\\
		heart  & \textbf{84.20$\pm$1.21} & 83.46$\pm$2.15\phantom{0} &81.98$\pm$0.99$\bullet$ & 76.79$\pm$0.49$\bullet$ & 80.99$\pm$0.99$\bullet$   & 74.07$\pm$0.00$\bullet$\\
		vehicle  & \textbf{87.48$\pm$1.07} & 83.54$\pm$1.68$\bullet$ &85.14$\pm$0.91$\bullet$ & 78.58$\pm$0.19$\bullet$ & 72.52$\pm$0.91$\bullet$   & 76.38$\pm$0.00$\bullet$\\
		sonar  & \textbf{92.06$\pm$1.42} & 88.89$\pm$1.42$\bullet$ &86.67$\pm$2.15$\bullet$ & 79.68$\pm$0.63$\bullet$ & 74.29$\pm$1.19$\bullet$   & 85.71$\pm$0.00$\bullet$\\
		glass  & \textbf{88.62$\pm$1.57} & 85.23$\pm$2.30$\bullet$ &86.46$\pm$1.15$\bullet$ & 81.85$\pm$1.15$\bullet$ & 78.46$\pm$0.00$\bullet$   & 86.15$\pm$0.00$\bullet$\\
		ionosphere  & 94.53$\pm$1.62 & 94.91$\pm$0.96\phantom{0} &91.32$\pm$1.10$\bullet$ & 94.15$\pm$0.38\phantom{0} & \textbf{95.28$\pm$0.00}\phantom{0}   & 94.34$\pm$0.00\phantom{0}\\
		phishing  & 96.90$\pm$0.14 & 96.77$\pm$0.12\phantom{0} &96.19$\pm$0.13$\bullet$ & \textbf{97.05$\pm$0.02}$\circ$ & 94.01$\pm$0.05$\bullet$   & 96.64$\pm$0.01$\bullet$\\
		\hline
		\multicolumn{2}{c|}{win/tie/lose} & \textbf{24/9/1} & \textbf{27/5/1} & \textbf{28/3/3} & \textbf{28/3/3} & \textbf{24/3/6}\\
		\hline
		\hline
		& NRS (ours) & MLP &ANT~\cite{ant:tanno:ICML19} &RSs& RFs & GBDTs \\
		\hline
		SARCOS &\phantom{0}\textbf{1.23$\pm$0.05} & \phantom{0}2.36$\pm$0.16 & 1.38* & 2.17$\pm$0.02 & \phantom{0}2.37$\pm$0.01 & \phantom{0}1.44$\pm$0.01\\
		\hline
	\end{tabular}
\end{table}

\textbf{Comparison among different algorithms:} Figure~\ref{fig:MLP_variants} shows that MLP-3 achieves comparable performance with MLP-2 and our NRS achieves higher accuracy than both MLP-2 and MLP-3 on all the 6 datasets. Notice that MLP-3 has even more parameters than our NRS and it demonstrates that it is the underlying neural random subspace method rather than the introduced more free parameters to achieve improved performance. In contrast to MLP-D which is inconsistent between training and inference by using a simple approximate average during inference, NRS matains the same structure during both stages and meanwhile achieves consistently better performance. Notice that MLP-D gets very poor results on satimage, letter, yeast and SARCOS which have low feature dimensions (c.f. Table~\ref{tab:ml-dataset-extensive}) and it indicates that using dropout at the input layer is not suitable for low-dimensional inputs. Also, MLP-2 outperforms MLP-D owing to BN, as pointed out in~\cite{batchnormalization:ioffe:ICLR15}.

Table~\ref{tab:ml-dataset-res} shows that NRS achieves the highest accuracy for the most of times in all classification datasets and the lowest mean square error (MSE) in the regression dataset compared with MLP, RSs, RFs, GBDTs, NDF~\cite{neural-decision-forests:kontschieder:ICCV15} and ANT~\cite{ant:tanno:ICML19}. As can be seen, our NRS method significantly outperforms MLP, NDF, RSs, RFs and GBDTs, since the win/tie/lose counts show that our NRS wins for most times and seldom loses and it demonstrate the effectiveness of NRS across datasets with various dimensionalities and sizes. Moreover, it is worth noting that although our method introduces randomness due to random permutations, it achieves a low standard deviation and is very robust, even more stable than MLP. 

\begin{table}[htbp]
	\small
	\caption{Model size (MB), total inference / training time (s) and accuracy (\%) comparison. We report the average results of 5 trials.}
	\label{tab:ml-dataset-speed-compare}
	\centering
	\setlength{\tabcolsep}{3.3pt}
	\renewcommand{\arraystretch}{0.5}
	\begin{tabular}{ll|c|c|r|c}
		\hline
		&\multirow{2}{*}{Method} & \multirow{2}{*}{Model size}& \multicolumn{2}{c|}{Time} & \multirow{2}{*}{Accuracy}  \\
		\cline{4-5}
		&&&Inference&Training&\\
		\hline
		\multirow{5}*{{GISETTE}}&NRS (ours) &35 & \oo0.17 & \phantom{0}62.51 & \textbf{97.82}  \\
		&NDF & 90.6 & \oo0.33&140.89 &97.10\\
		&RSs & 6.6 & \oo1.87  & \phantom{0}57.83 & 95.60\\
		&RFs &3.6 & \oo0.12 & \textbf{\phantom{0}\phantom{0}0.67} &96.70 \\
		&ThunderGBM RFs & 0.6 & \oo2.77 & \phantom{0}24.96 & 93.60 \\
		&GBDTs &\textbf{0.2} & \textbf{\oo0.01} & 181.14 & 96.70 \\
		&ThunderGBM GBDTs & 0.6 & \oo2.04 & \phantom{0}18.78 & 91.79 \\
		\hline
		\multirow{5}*{{MNIST}}&NRS (ours) & 9.6 & \textbf{\oo0.17} & 194.45 & \textbf{98.42}  \\
		&NDF &  83.6 & \oo2.63& 1263.19 & 98.29\\
		&RSs & 115 & \oo5.22 & 196.12 & 96.65 \\
		&RFs &137 & \oo0.31 & \textbf{2.09} &96.85 \\
		&ThunderGBM RFs & 6.1 & \oo0.76 & 19.43 & 93.16 \\
		&GBDTs & \textbf{1.7} & \oo0.42 & 2877.78 & 94.87 \\
		&ThunderGBM GBDTs & 6.1 & \oo0.99 & 23.66 & 93.78 \\
		\hline
		\multirow{5}*{{letter}}&NRS (ours) & \textbf{3.4} & \textbf{\oo0.19} & 43.49 & \textbf{97.78}  \\
		&NDF & 85.3 & \oo2.96& 1127.07 & 95.94\\
		&RSs & 151 & \oo5.03 & 4.95 & 96.50\\
		&RFs & 106 & \oo0.39& \textbf{0.38} &96.12 \\
		&ThunderGBM RFs & 15.9 & \oo0.35 & 16.03 & 93.29 \\
		&GBDTs &4.5 & \oo0.27 & 50.88 & 92.04 \\
		&ThunderGBM GBDTs & 15.9 & \oo0.33 & 15.20 & 92.99 \\
		\hline
		\multirow{5}*{{SUSY}}&NRS (ours) & \textbf{0.5} & 14.64 & 103.26 & \textbf{80.37}  \\
		&NDF & 2.2  &26.74 & 4134.57 & 78.31\\
		&RSs & 4596 & 95.86  &  3621.16 &  79.87\\
		&RFs & 5711  &39.25  & 1651.36  & 80.02 \\
		&ThunderGBM RFs & 0.6  & \textbf{\oo3.86}  &32.93  & 80.23 \\
		&GBDTs & 1.2 & \oo5.53 & 7361.88  & 80.26  \\
		&ThunderGBM GBDTs &  0.6 & \oo3.98  & \textbf{31.33}  & 80.25  \\
		
		\hline
	\end{tabular}
\end{table}

In Tabel~\ref{tab:ml-dataset-speed-compare} we compare the speed and size of NRS with RSs, RFs and GBDTs. Note that although we reduce the number of trees for RFs from 500 to 100 on MNIST, the accuracy drops slightly (from 96.96\% to 96.85\%) while the model size is reduced by 5 times (from 680M to 137M). Table~\ref{tab:ml-dataset-speed-compare} shows that GPU-based methods ThunderGBM and our NRS greatly accelerate the training process on the large-scale dataset SUSY, yielding much higher efficiency when compared to conventional forest methods. NRS achieves faster training and inference speed and smaller model size than NDF, which demonstrates that our NRS makes better use of deep learning platforms support and achieves higher efficiency than NDF by using existing CNN layers for the implementation. Compared to these forest methods, NRS achieves the highest accuracy and the fastest inference speed on MNIST and letter, and also the smallest model size on letter and SUSY. NRS achieves the highest accuracy on GISETTE but the model size is larger than other forest methods, indicating that NRS may be unfriendly to those datasets with non-sparse high dimensionalities in terms of model size. Note that although we use smaller $nMul$ values in Table~\ref{tab:ml-dataset-speed-compare} than the experiments reported in Table~\ref{tab:ml-dataset-res}, NRS's accuracy in Table~\ref{tab:ml-dataset-speed-compare} are still similar to those in Table~\ref{tab:ml-dataset-res} (e.g., 97.85 in Table~\ref{tab:ml-dataset-res} vs. 97.78 in Table~\ref{tab:ml-dataset-speed-compare} on letter). Effect of the sensitivity of NRS's hyperparameters such as $nMul$ will be studied in Sec~\ref{sec:ablation-study}.  

\subsubsection{Hyperparameters studies} \label{sec:ablation-study}
We choose the 4 \emph{largest} datasets among the first 6 machine learning datasets, i.e., GISETTE, MNIST, letter and USPS to study the sensitivity of hyperparameters in our method NRS. Hyperparameters studies include three parts: expansion rate $nMul$, number of channels per group $nPer$, expansion height/width $nH$ and the number of group convolution layers.

\begin{figure}
	\centering
	\subfloat[Classification results using different expansion rates $nMul$.]{
		\label{fig:nMul-ablation}
		\begin{minipage}[b]{\linewidth}
			\includegraphics[width=0.24\linewidth]{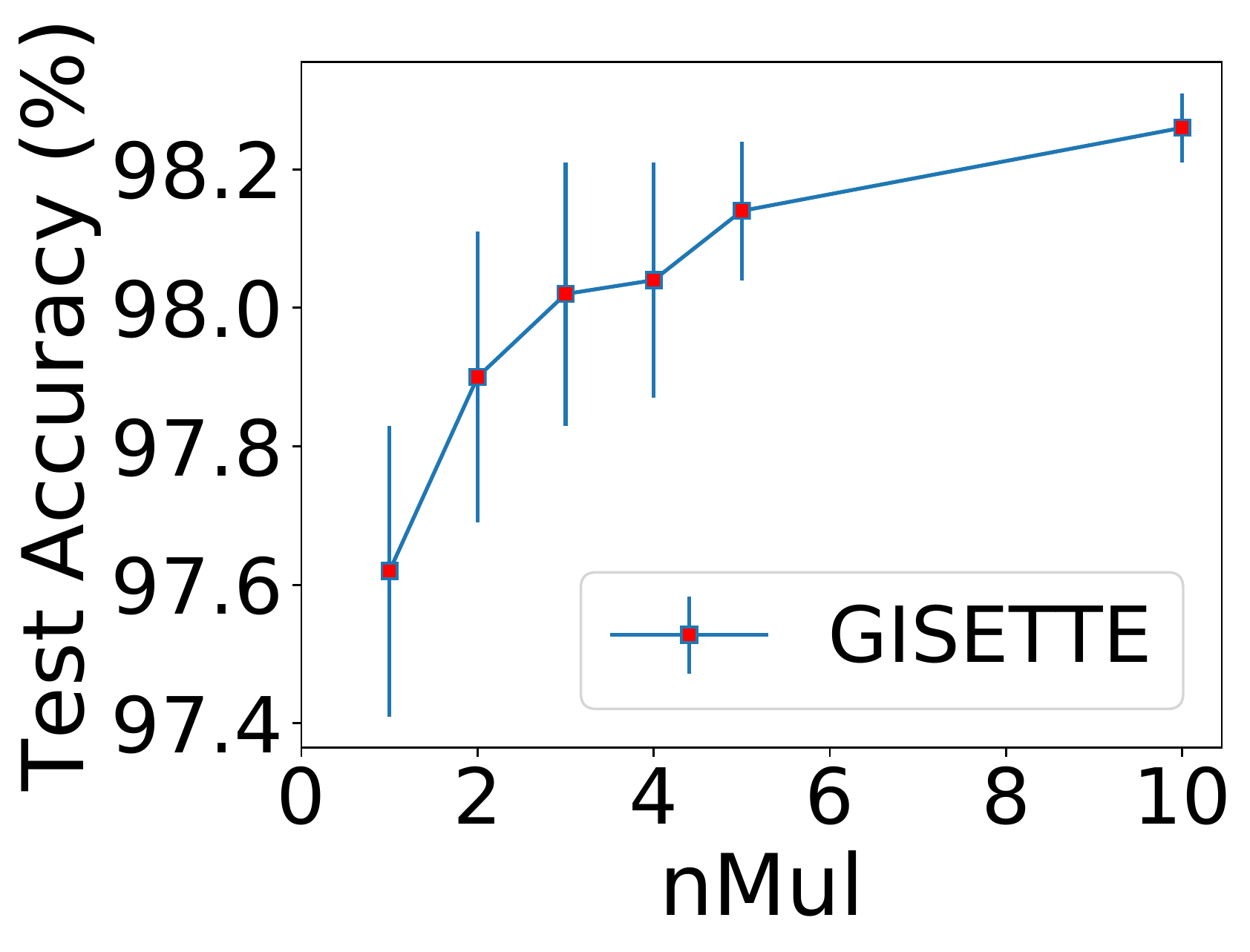}
			\includegraphics[width=0.24\linewidth]{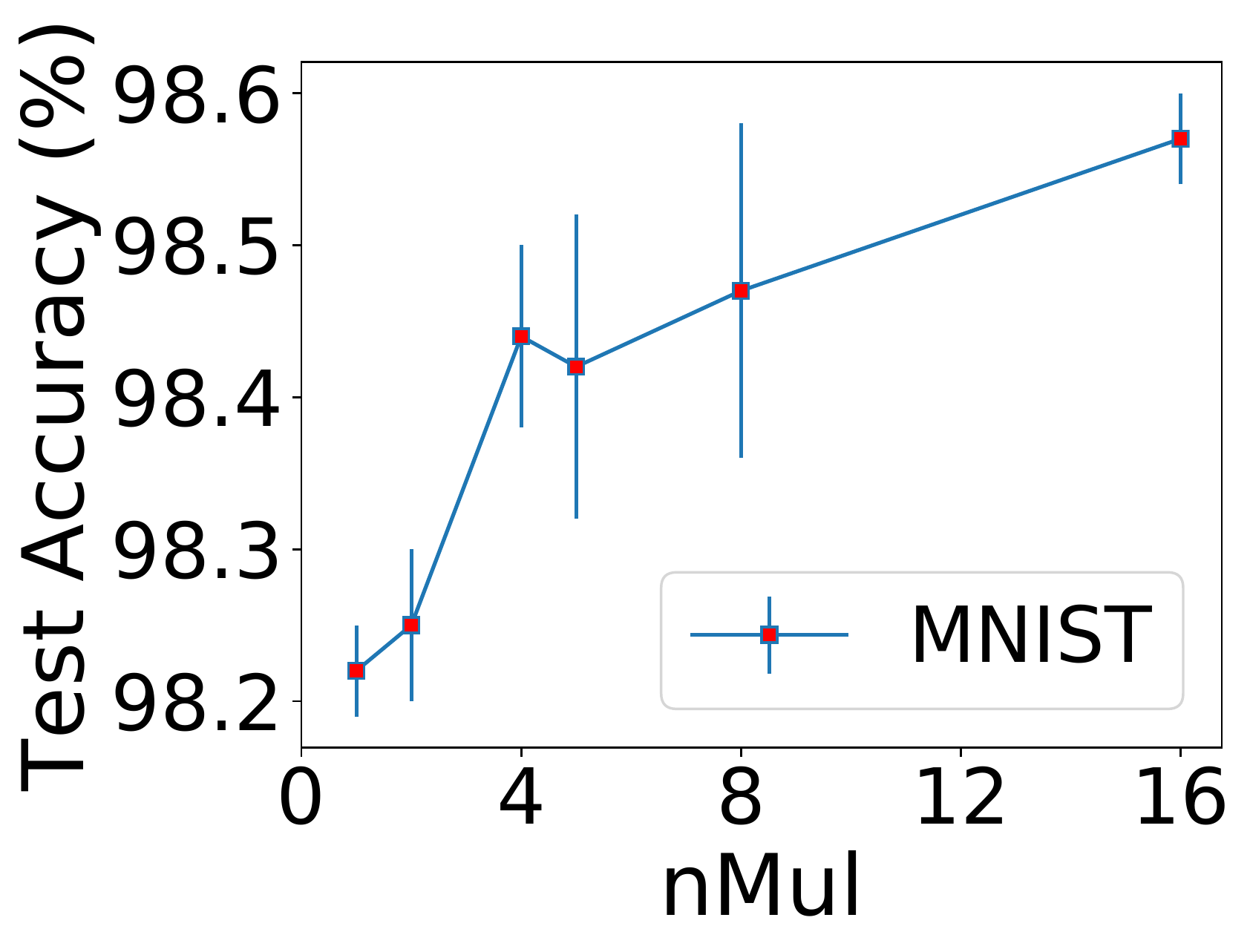}
			\includegraphics[width=0.24\linewidth]{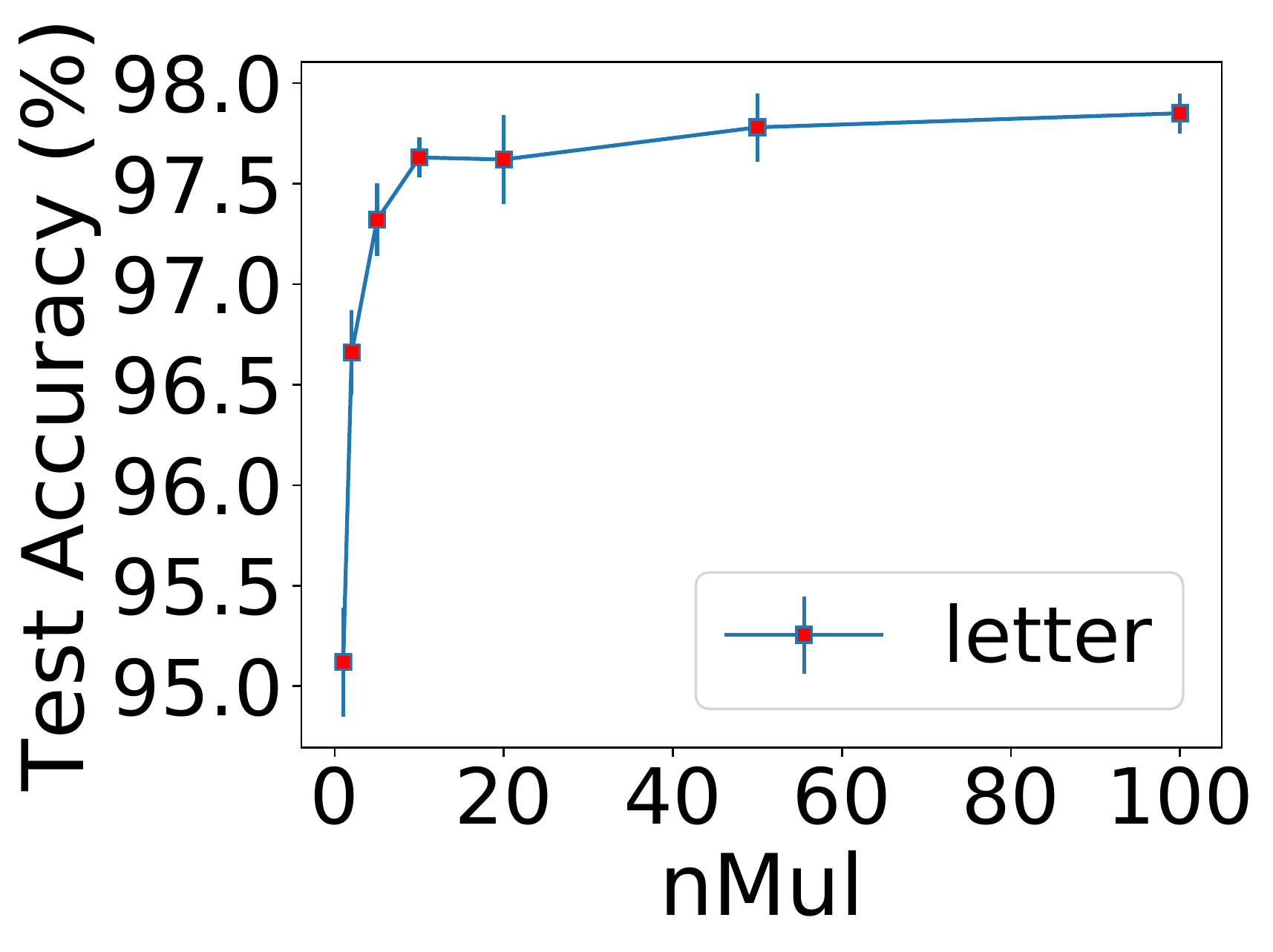}
			\includegraphics[width=0.24\linewidth]{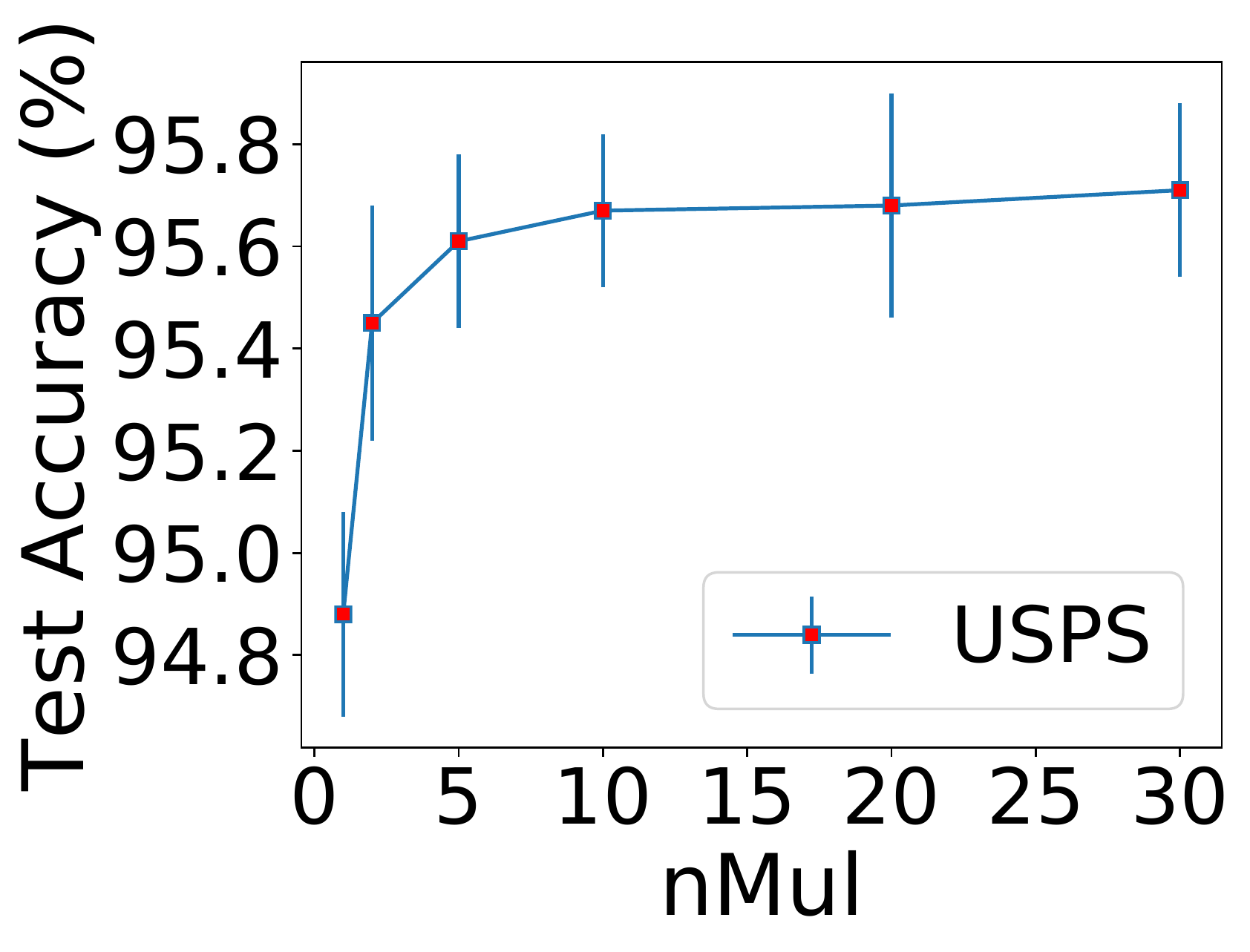}
	\end{minipage}}
	\\
	\subfloat[Classification results using different number of channels per group $nPer$.]{
		\label{fig:nPer-ablation}
		\begin{minipage}[b]{\linewidth}
			\includegraphics[width=0.24\linewidth]{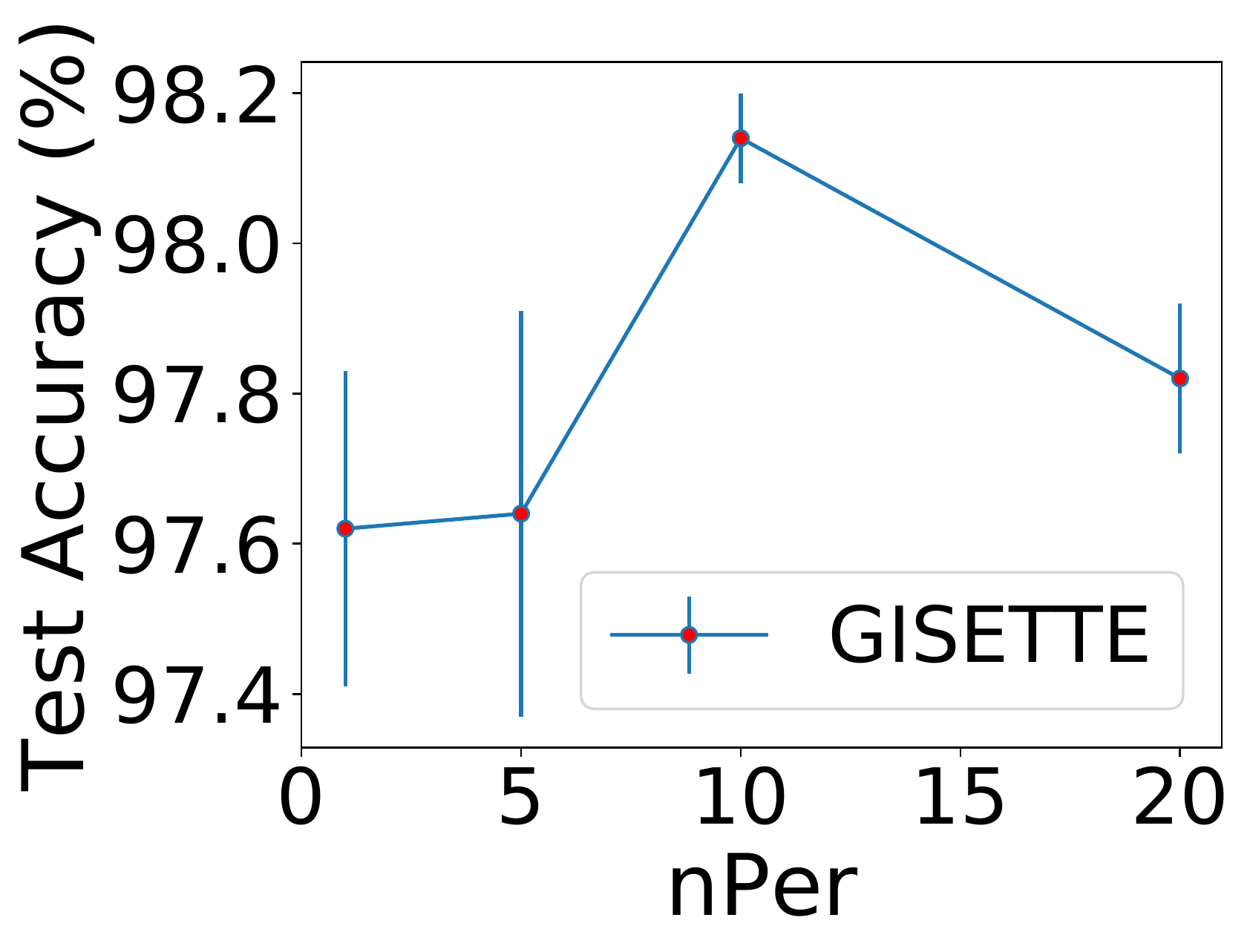}
			\includegraphics[width=0.24\linewidth]{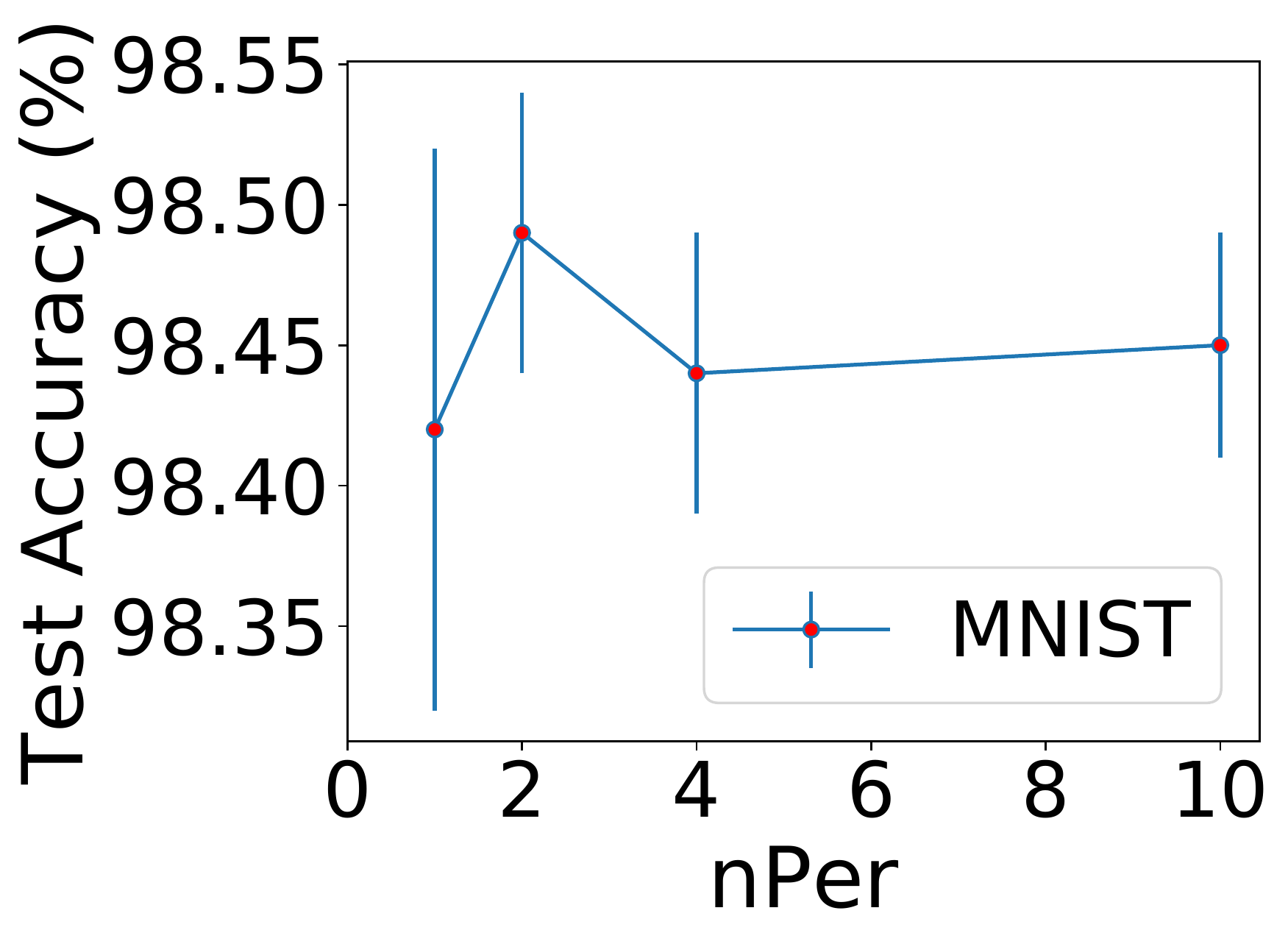}
			\includegraphics[width=0.24\linewidth]{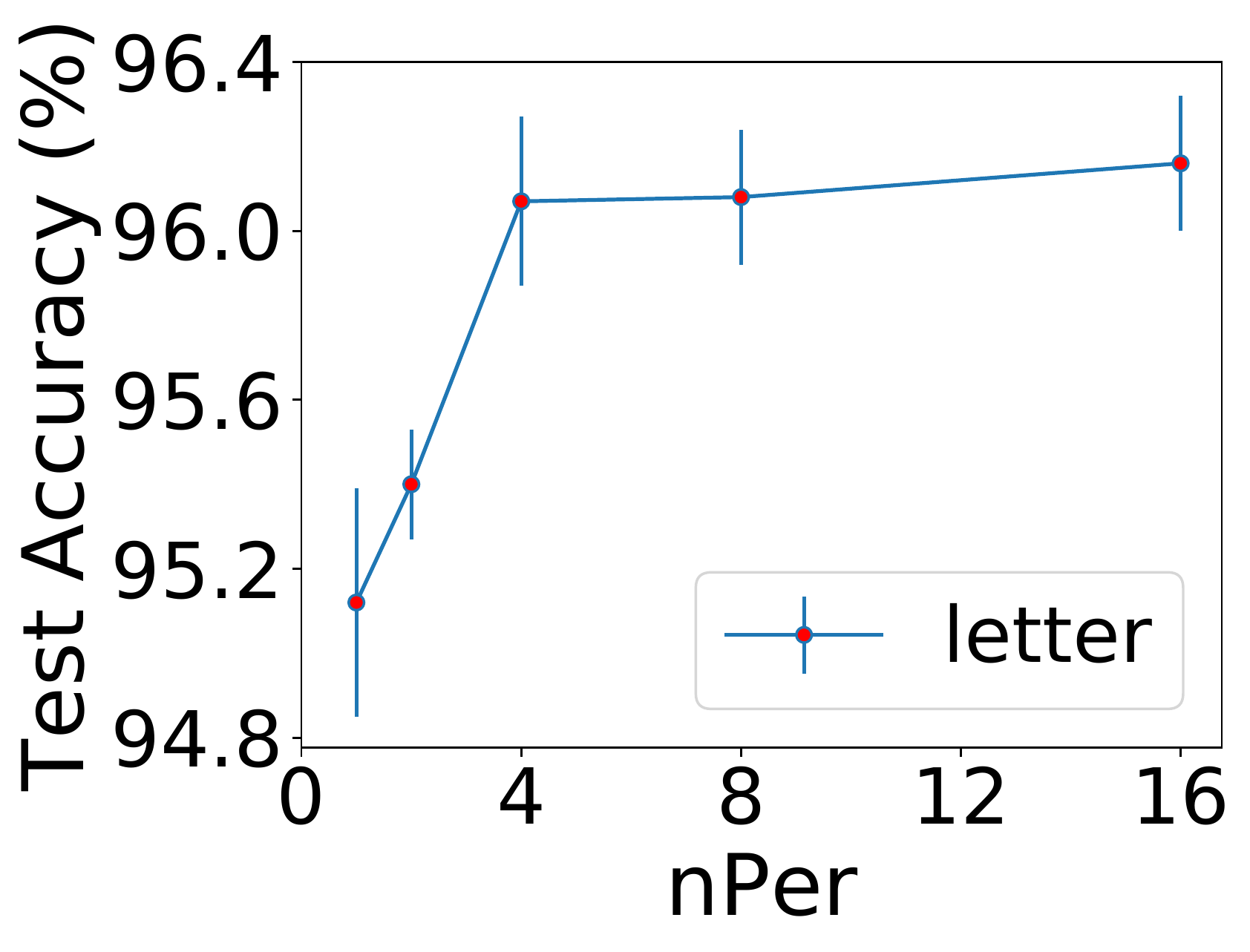}
			\includegraphics[width=0.24\linewidth]{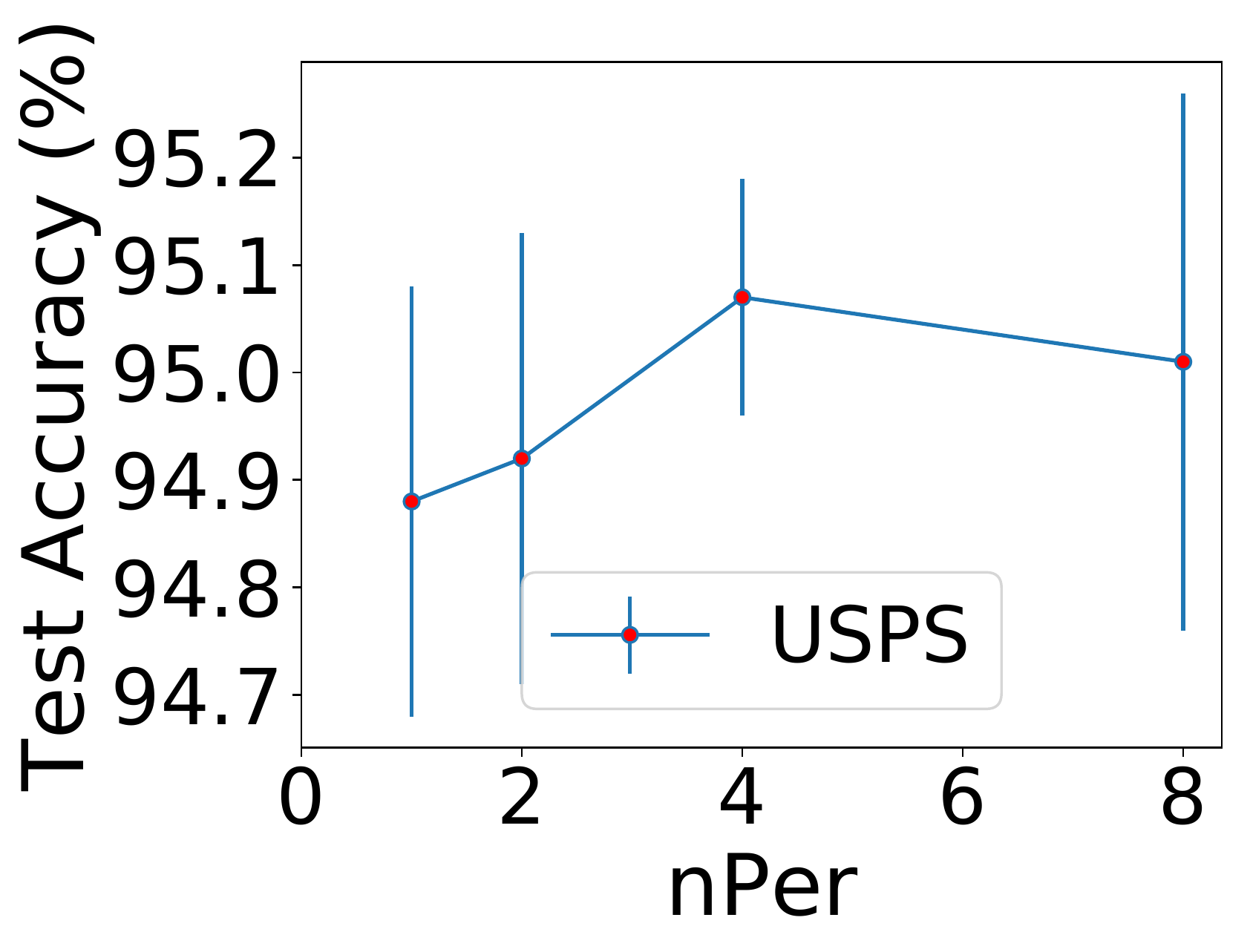}
	\end{minipage}}
	\\
	\subfloat[Classification results using different expansion height/width` $nH$.]{
		\label{fig:dH/dW-ablation}
		\begin{minipage}[b]{\linewidth}
			\includegraphics[width=0.24\linewidth]{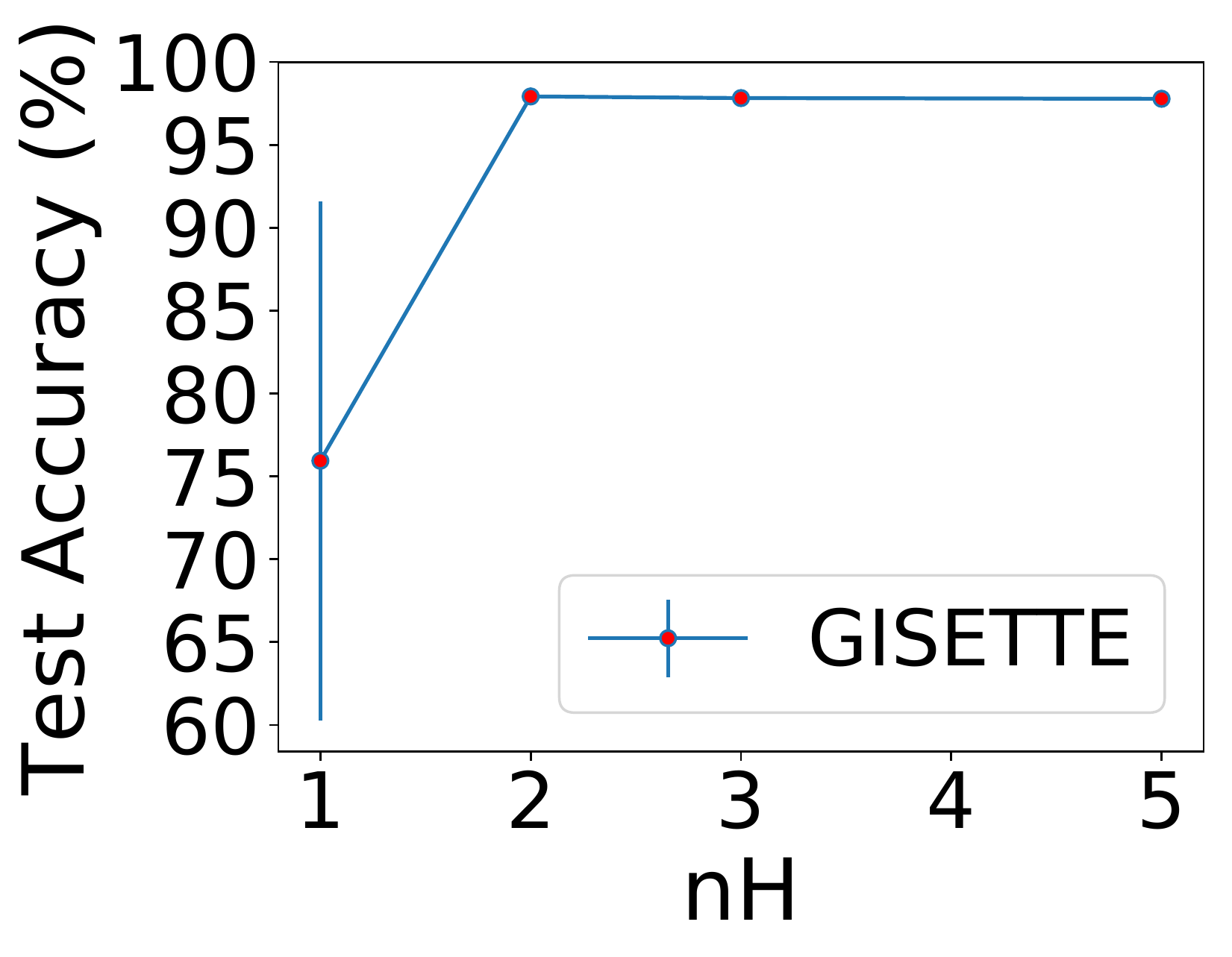}
			\includegraphics[width=0.24\linewidth]{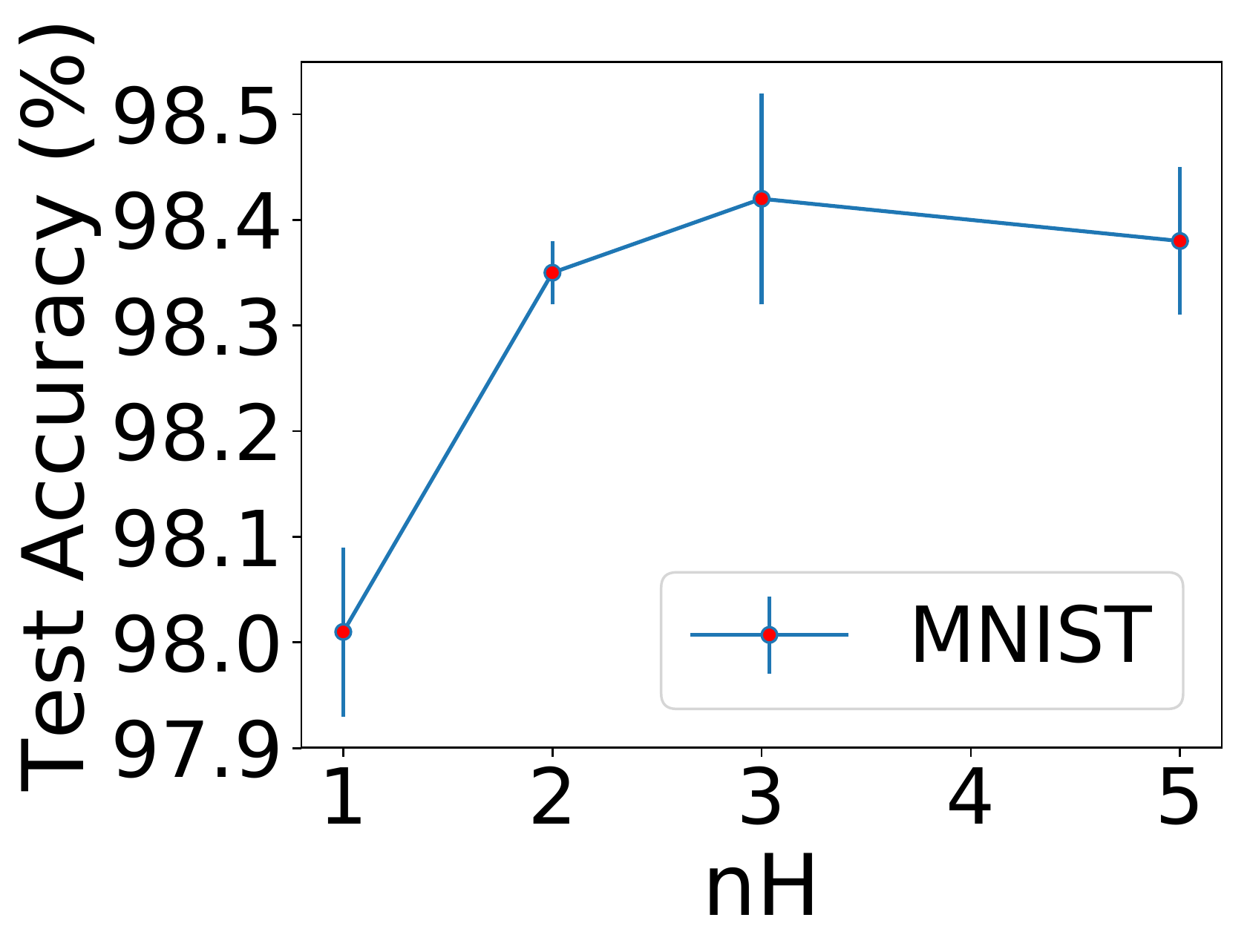}
			\includegraphics[width=0.24\linewidth]{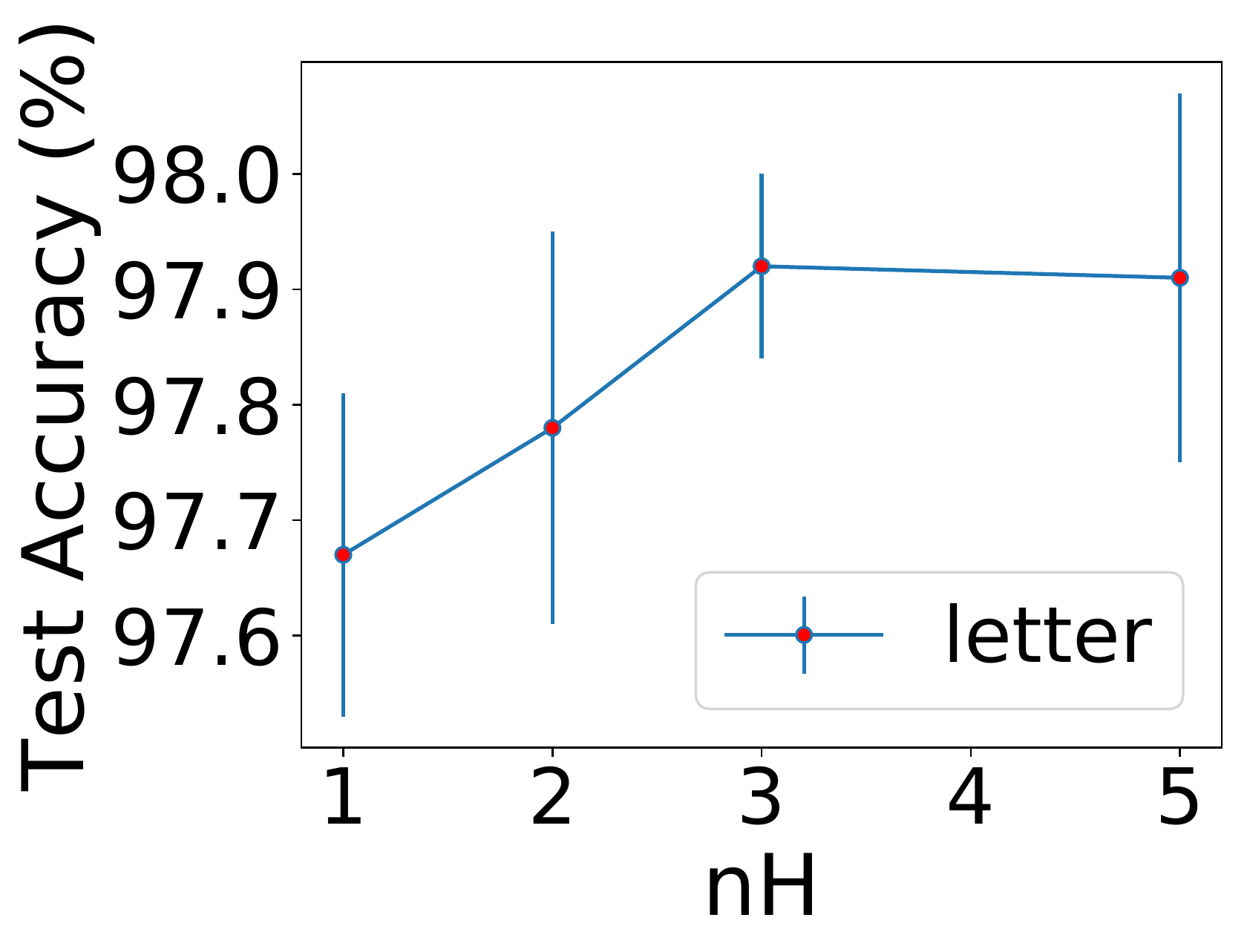}
			\includegraphics[width=0.24\linewidth]{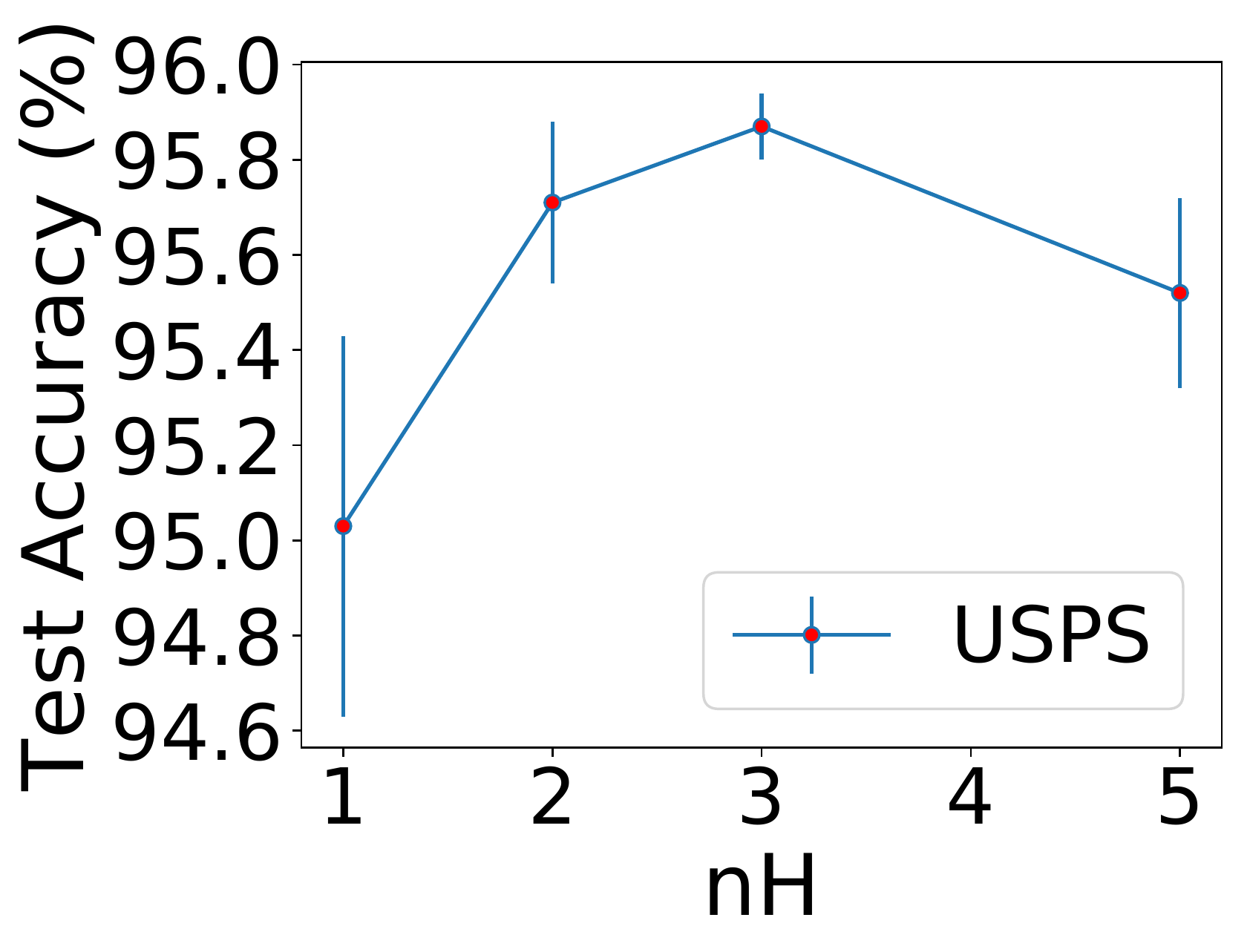}
	\end{minipage}}
	\caption{Hyperparameters studies of $nMul$, $nPer$ and $nH$ on GISETTE, MNIST, letter and USPS (from left to right in each sub figure). We plot the average accuracy and standard deviation of 5 trails at each point.}
\end{figure}

\textbf{Expansion rate.}  As is known in RSs, RFs and GBDTs, we can increase the number of decision trees to boost performance. Similarly, we can increase $nMul$
in NRS to increase the number of trees in our ensemble and we conduct studies about $nMul$. Here we keep other settings the same as before for all experiments. The results in Figure~\ref{fig:nMul-ablation} show that when $nMul$ grows, the average accuracy increases and the standard deviation becomes smaller. It indicates that as $nMul$ grows, more trees (random subspaces) are integrated into our model and the performance becomes better and our model gets more robust. 

\textbf{Number of channels per group.}  We can also increase $nPer$ to increase the number of features utilized in each random subspace. Here we set $nMul$ to 1 for all experiments and other settings remain the same. The results in Figure~\ref{fig:nPer-ablation} show that when $nPer$ grows, the test accuracy will increase at first and then it will become stable or slightly decrease. It means that as $nPer$ increases, the capacity of each random subspace and hence the whole model will also increase, thus the accuracy will also increase at first. However, the model is more likely to overfit with large $nPer$ and model capacity and the performance will not continue to improve. Also, notice that when we increase $n
Per$ to equal the number of channels, group convolution becomes standard convolution and each filter will learn from all features. It is obviously a departure from the random subspace idea which requires each base learner to learn from different feature subsets and the performance also becomes worse correspondingly. Hence, it further verifies the choice of group convolution instead of standard convolution and it also indicates that we can get better results in Table~\ref{tab:ml-dataset-res} by choosing appropriate $nPer$.

\textbf{Expansion height/width.} We set $nMul$ to 1, 5, 50 and 20 for GISETTE, MNIST, letter and USPS and we set $nPer$ to 1 for all these datasets. The results in Figure~\ref{fig:dH/dW-ablation} show that when $nH$ is very small, i.e., equals 1, the result is bad, especially for GISETTE. When $dH/dW$ grows, the result becomes better and will not continue to improve when it grows beyond 3. Therefore, 2 or 3 is a good choice for $nH$ in terms of accuracy and efficiency and we use $nH=3$ in all our experiments in this paper for simplicity. 

\textbf{Number of group convolution layers.} We also study the effect of more group convolution layers and we compare the results of NRS with 1 group convolution layer and 2 group convolution layers. Experimental results show that both settings achieve comparable results on these 4 datasets while more group convolution layers brings more computing overhead. Hence, we use NRS with 1 group convolution layer in all our experiments in this paper.

\subsubsection{Ablation Study}

\begin{table}[t]
	\caption{Classification results evaluated on satimage, GISETTE, MNIST and letter. Starting from our baseline, we gradually add sparse initialization and freezed weight scheme in the FC layer (equivalently implementation of random permutation) in our NRS for ablation studies. }
	\label{tab:ablation}
	\small
	\centering
	\setlength{\tabcolsep}{1.0pt}
	\renewcommand{\arraystretch}{0.7}
	\renewcommand{\multirowsetup}{\centering}
	\begin{tabular}{c|c|c|c|c|c|c}
		\hline
		 &\multicolumn{2}{c|}{Scheme} & \multicolumn{4}{c}{Accuracy}\\
		\cline{2-7}
		&sparse init. & freezed weight& MNIST & satimage  & german& heart \\
		\hline
		(a) &$\times$ &$\times$&98.21$\pm$0.10&91.01$\pm$0.42& 72.20$\pm$0.78&80.99$\pm$0.60\\
		(b) & \checkmark& $\times$&98.18$\pm$0.08&91.06$\pm$0.16&71.87$\pm$1.05&81.73$\pm$1.44\\
		(c) &$\times$ &\checkmark&98.03$\pm$0.05&91.08$\pm$0.49&74.60$\pm$1.06&82.96$\pm$0.92\\
		(d) &\checkmark&\checkmark&\textbf{98.57$\pm$0.03}&\textbf{91.52$\pm$0.31}&\textbf{76.60$\pm$0.53}&\textbf{84.20$\pm$1.21}\\
		\hline
	\end{tabular}
\end{table}

Notice that the random permutation operation is a linear operation and can be equivalently implemented by a FC layer with sparse weight matrix and we conduct ablation studies on this special FC layer. This FC layer has weight matrix $W^{FC} \in \mathbb{R}^{d\times{Md}}$, which maps input $\boldsymbol{x}\in \mathbb{R}^d$ to $\boldsymbol{y}\in \mathbb{R}^{Md}$, where $M=nH\times{nH}\times{nMul}$ defined as before. Then we have $y_j = x_i$, where $i=\boldsymbol{\sigma}^{\lfloor{j/M}\rfloor}_{j\text{ mod } M}$. Correspondingly, the weight matrix is highly sparse with $W^{FC}_{ij}=1$ $(j=1,\cdots, M{d})$ and other weights are set to 0, i.e., each output unit is connected to one particular input unit with weight 1. Hence, our random permutation operation can be considered as a FC layer with two special properties: (i) sparse initialization: the initial weight is highly sparse and specified by the generated permutation. (ii) freezed weight: the weight matrix of this layer is fixed during training. To further validate the effectiveness of our random permutation operation in NRS, we implement the random permutation by a FC layer and ablate its two special properties.

We choose 1 large dataset MNIST, 1 medium dataset satimage and 2 small datasets german and heart. Table~\ref{tab:ablation} shows that the row (d) achieves superior performance over other strategies consistently on all the 4 datasets and it shows that sparse initialization and freezed weight properties in the random permutation has its effectiveness. Also, it is worth noting that the row (c) achieves higher accuracy than both (a) and (b) on the 2 small datasets german and heart, which indicates that the over-parameterization in the FC layer makes the model more likely to overfit and the weight freezing strategy is beneficial, especially on small datasets. Moreover, contrary to the implementation by a FC layer, the random permutation operation has no parameters and small FLOPs and hence it is much more efficient than a FC layer. In conclusion, the random permutation operation in NRS achieves both higher accuracy and efficiency than a FC layer implementation.

\subsection{Document Retrieval Datasets}\label{sec:retrieval}
Aside from the machine learning datasets used before, we apply NRS into the challenging document retrieval task, which also has vectorized inputs. 


\textbf{Evaluation metrics: } We use the standard ranking accuracy metric to evaluate the rank functions generated by learning to rank algorithms: normalized discount cumulative gain (NDCG$@k$)~\cite{ndcg02Kalervo}. It has been proposed to handle multiple levels of relevance:
\begin{equation}
    \text{NDCG}@k = Z_k \sum_{j=1}^k \left\{
\begin{aligned}
2^{rel(j)}-1 &\quad \text{if} \ j=1 \\
\frac{2^{rel(j)}-1}{log(j)} &\quad \text{if} \ j>1
\end{aligned}
\right.,
\end{equation}
where $rel(j)$ is the integer rating of the $j$-th document, and the normalization constant $Z_k$ is chosen such that the perfect list gets a NDCG score of 1.

\textbf{Implementation details: }
In our experiments, we used two widely used benchmark datasets, Microsoft 10K and Microsoft 30K~\cite{mslrweb}. Each query-document pair is represented with a 136-dimensional feature vector. The groundtruth is a multiple-level relevance judgment, which takes five values from 0 (irrelevant) to 4 (perfectly relevant). We report the results with different NDCG cutoff values 1, 5 and 10 to show the performance of each method at different positions. 

In our experiments, 4 typical listwise ranking methods ListNet~\cite{listnet:zhe:ICML07}, ApproxNDCG~\cite{apxndcg:tao:IPM10}, RankCosine~\cite{rankcosine:tao:IPM08}, WassRank~\cite{wassrank:yu:WSDM19} as well as 1 pariwise ranking method RankNet~\cite{ranknet:burges:ICML05} are used as our baselines. Following prior work \cite{listnet:zhe:ICML07, wassrank:yu:WSDM19}, a simple 1-layer feed-forward neural network (a dropout rate of 0.01) with the Sigmoid activation function is used as the ranking function. For NRS, we set $nMul$ to 2, $nPer$ to 1 and $nH$ to 3 in all our experiments in this section. Following \cite{wassrank:yu:WSDM19}, we used the L2 regularization with a decaying rate of 1e-3 and the Adam~\cite{adam} optimizer with a learning rate of 1e-3. In particular, each dataset has been randomly partitioned into five equal sized subsets. In each fold, three subsets are used as the training data, the remaining two subsets are used as the validation data and the testing data, respectively. We use the training data to learn the ranking model, use the validation data to select the hyper parameters based on NDCG@10, and use the testing data for evaluation. Finally, we report the ranking performance based on the averaged evaluation scores across five folds.

\begin{table}[t]
	\caption{Performance of different models on Microsoft 10K and Microsoft 30K. The best result of each setting is indicated in bold. The best result overall is indicated in bold and marked with $*$. $\blacktriangle$ denotes	our method.}
	\label{tab:MSLR-WEB10k}
	\centering
	\small
	\setlength{\tabcolsep}{1.0pt}
	\renewcommand{\arraystretch}{0.65}
	\renewcommand{\multirowsetup}{\centering}
	\begin{tabular}{c|c|c|c|c|c|c}
		\hline
		\multirow{2}{*}{Method} &\multicolumn{3}{c|}{Microsoft 10K} &\multicolumn{3}{c}{Microsoft 30K}\\
		\cline{2-7}
		&NDCG@1 & NDCG@5 & NDCG@10 & NDCG@1 & NDCG@5 & NDCG@10 \\
		\hline
		RankNet~\cite{ranknet:burges:ICML05}&0.306\oo&0.341\oo&0.370\oo&0.300\oo&0.339\oo&0.369\oo\\ 
		RankNet+NRS $\blacktriangle$&\textbf{0.370}\oo&\textbf{0.378}\oo&\textbf{0.402}\oo&\textbf{0.363}\oo&\textbf{0.377}\oo&\textbf{0.400}\oo\\ 
		\hline
		ListNet~\cite{listnet:zhe:ICML07}&0.348\oo&0.360\oo&0.381\oo&0.393\oo&0.398\oo&0.419\oo\\ 
		ListNet+NRS $\blacktriangle$&\textbf{0.388}\oo&\textbf{0.392}\oo&\textbf{0.414}\oo&\textbf{0.403}\oo&\textbf{0.402}\oo&\textbf{0.424}\oo\\
		\hline
		ApxNDCG~\cite{apxndcg:tao:IPM10}&0.133\oo&0.166\oo&0.196\oo&0.118\oo&0.159\oo&0.192\oo\\ 
		ApxNDCG+NRS $\blacktriangle$&\textbf{0.399}\oo&\textbf{0.397}\oo&\textbf{0.418}\oo&\textbf{0.407}\oo&\textbf{0.403}\oo&\textbf{0.424}\oo\\
		\hline
		RankCosine~\cite{rankcosine:tao:IPM08}&0.380\oo&\textbf{0.397}\oo&0.417\oo&0.386\oo&0.401\oo&0.422\oo\\ 
		RankCosine+NRS $\blacktriangle$&\textbf{0.399}\oo&0.396\oo&\textbf{0.418}\oo&\textbf{0.412}\oo&\textbf{0.408}\oo&\textbf{0.429}\oo\\
		\hline
		WassRank~\cite{wassrank:yu:WSDM19}&0.255\oo&0.272\oo&0.195\oo&0.404\oo&0.398\oo&0.418\oo\\ 
		WassRank+NRS $\blacktriangle$&\textbf{0.392}\oo&\textbf{0.393}\oo&\textbf{0.414}\oo&\textbf{0.424}\oo&\textbf{0.416}\oo&\textbf{0.433}\oo\\
		\hline
		uRank~\cite{urank:zhu:WSDM20} & 0.458\oo & 0.443\oo & 0.458\oo & 0.465\oo & 0.453\oo & 0.471\oo\\
		uMart~\cite{urank:zhu:WSDM20} & \textbf{0.481*} & \textbf{0.474*} & \textbf{0.495*} & \textbf{0.505*} & \textbf{0.494*} & \textbf{0.512*}\\
		\hline
	\end{tabular}
\end{table}

\textbf{Comparison among different algorithms: }As can be seen from Table~\ref{tab:MSLR-WEB10k}. NRS achieves consistent improvements under various baseline ranking models on both Microsoft 10K and Microsoft 30K. The experimental results demonstrate that the integration of NRS into various baseline ranking models achieve superior performance consistently, which indicates that NRS has the potential to improve performance when combined with different algorithms on document retrieval tasks. We also compare with the state-of-the-art algorithms uRank~\cite{urank:zhu:WSDM20} and uMart~\cite{urank:zhu:WSDM20}. Although there is a gap between our results and the state-of-the-art results , we aim to show that NRS has the potential to further boost performance when combined with emerging state-of-the-art algorithms, e.g., uRank.

\subsection{Computer Vision Datasets}
We then  move from vectorized inputs to image data and we evaluate NRS in CNN architectures for both 2D image recognition tasks in this section and 3D recognition tasks in the next section. For image recognition tasks, NRS is used after GAP to non-linearly transform the GAP output vector at the end of the network and we evaluate it on both fine-grained visual categorization tasks in Sec~\ref{sec:CUB200-2011} and the large-scale ImageNet ILSVRC-12 in Sec~\ref{sec:ImageNet}. Moreover, we demonstrate that NRS can further be installed across all layers in CNNs (e.g., SENet~\cite{senet:hujie:arxiv}) and we evaluate it on CIFAR-10, CIFAR-100 and ImageNet ILSVRC-12 in Sec~\ref{sec:senet}. 

\subsubsection{Fine-grained Visual Categorization} \label{sec:CUB200-2011}
We then evaluate NRS in CNN architectures for image recognition. NRS is used after GAP to non-linearly transform the GAP output vector at the end of the network. First, this section evaluates NRS with ResNet-50~\cite{resnet:he:CVPR16} and VGG-16~\cite{vgg:simonyan:ICLR15} on the Birds, Aircraft and Cars datasets. We compare our method with baseline models and one representative higher-order pooling method.

\begin{table}[t]\small
	\caption{Comparison of representation dimensions, parameters, FLOPs, inference time per image (ms) and accuracy (\%) on fine-grained benchmarks. The inference time is recorded with batch size of 1 on CPU and GPU. $\blacktriangle$ denotes our method.}
	\label{tab:cub200-result}
	\centering
	\setlength{\tabcolsep}{0.6pt}
	\renewcommand{\arraystretch}{0.9}
	\renewcommand{\multirowsetup}{\centering}
	\begin{tabular}{l|r|c|c|c|c|c|c|c}
		\hline
		\multirow{2}{*}{Method}      & \multirow{2}{*}{\,\#Dim\,} & \multirow{2}{*}{\,\#Params\,} &\multirow{2}{*}{\#FLOPs} & \multicolumn{2}{c|}{\,Inference Time\,} & \multicolumn{3}{c}{Accuracy}\\
		\cline{5-9}
		&&&& CPU & GPU & \,Bird\, & \,Aero\, & \,Car\, \\
		\hline
		ResNet-50&  \phantom{0}\phantom{0}2K&\oo23.92M&\oo16527.59M & \oo540.48&28.16&84.0&88.6&89.2\\ 
		CBP~\cite{cbp:gao:CVPR2016}  & 14K &\oo25.15M & \oo23212.04M  & 1594.64 & 46.19 &81.6 &81.6 &88.6 \\ 
		iSQRT~\cite{isqrt:li:CVPR2018}  &32K&443.14M& 121345.61M &2094.78  & 97.05 & \textbf{88.1}& 90.0&92.8 \\ 
		ResNet-50+NRS $\blacktriangle$& \phantom{0}\phantom{0}4K&\oo26.70M&\oo16530.79M& \oo541.57&28.38& 86.7&\textbf{92.8}&\textbf{93.4}\\ 
		\hline
		VGG-16& \phantom{0}0.5K&\oo15.34M&\oo61441.02M & \oo644.18&28.24&78.7&82.7&83.7\\ 
		B-CNN~\cite{bcnn:lin:ICCV15}&262K&\oo67.14M&\oo61751.10M & \oo856.46&31.90&84.0&84.1&90.6\\ 
		CBP~\cite{cbp:gao:CVPR2016}  &8K & \oo16.35M & \oo68046.35M   & 1116.38 & 41.27 &84.3 &84.1 &91.2 \\
		iSQRT~\cite{isqrt:li:CVPR2018}  &32K&\oo40.98M& \oo61493.24M & \oo770.52 & 30.33 &\textbf{87.2} &\textbf{90.0} &\textbf{92.5} \\ 
		VGG-16+NRS $\blacktriangle$& \phantom{0}\phantom{0}3K&\oo17.11M&\oo61443.83M & \oo645.19&28.58 & 84.4 & 89.6 & 91.5\\  
		\hline
	\end{tabular}
\end{table}

\textbf{Implementation details:} For fair comparisons, we follow~\cite{bcnn:lin:ICCV15} for experimental setting and evaluation protocal. We crop $448\times448$ patches as input images for all datasets. For baseline models, we replace the 1000-way softmax layer of ResNet-50 pretrained on ImageNet ILSVRC-12~\cite{ILSVRC2012:russakovsky:IJCV15} with a $k$-way softmax layer for finetuing, where $k$ is the number of classes in the fine-grained dataset. We replace all FC layers of pretrained VGG-16 with a GAP layer plus a $k$-way softmax layer to fit $448\times448$ input. We fine tune all the networks using SGD with batch size of 32, a momentum of 0.9 and a weight decay of 0.0001. We train the networks for 65 epochs, initializing the learning rate to 0.002 which is devided by 10 every 20 epochs. For NRS models, we replace the 1000-way softmax layer of pretrained ResNet-50 with our NRS module, specifically, random permutations, 1 group convolution layer and a $k$-way softmax layer ($k$ is number of classes), which is called ResNet-50+NRS. Here we set $nMul$, $nPer$ and $nH$ to 2, 64 and 3, respectively. Moreover, we also use the pretrained VGG-16 as our backbone network to construct VGG-16+NRS in a similar way, except that we set $nMul$ to 6 considering different feature dimensionalities. We fine tune our models under the same setting as the baseline models. These models integrating NRS are trained end-to-end as the baseline models.

\textbf{Comparison among different algorithms:} Table~\ref{tab:cub200-result} shows that our NRS method achieves significant improvements compared to baseline models, with negligible increase in parameters, FLOPs and real running time. It is worth mentioning that VGG-16+NRS achieves $7.2\%, 8.3\%$ and $9.3\%$ relative improvement over baseline models on Birds, Aero (Aircraft) and Cars, respectively. Besides, our NRS performs consistently better than B-CNN~\cite{bcnn:lin:ICCV15} on all the 3 datasets under the VGG-16 architecture despite using much fewer parameters, FLOPs and real running time. CBP~\cite{cbp:gao:CVPR2016} greatly reduces the number of parameters at the cost of much more FLOPs and real running time. Although iSQRT~\cite{isqrt:li:CVPR2018} achieves higher accuracy than our NRS in most cases, it brings heavy computational overhead when compared with our NRS, especially for ResNet-50. In short, compared with other non-linear methods, NRS achieves a good tradeoff between accuracy and efficiency. Furthermore, the learning curves in Figure~\ref{fig:loss-accuracy} shows that NRS can greatly accelerate the convergence and achieves better results both in accuracy and convergence speed than baseline methods (the red curves vs. the green curves). It indicates that NRS can effectively learn non-linear feature representations and achieves good results on fine-grained recognition.


\begin{figure}
    \centering
    \includegraphics[width=0.47\linewidth]{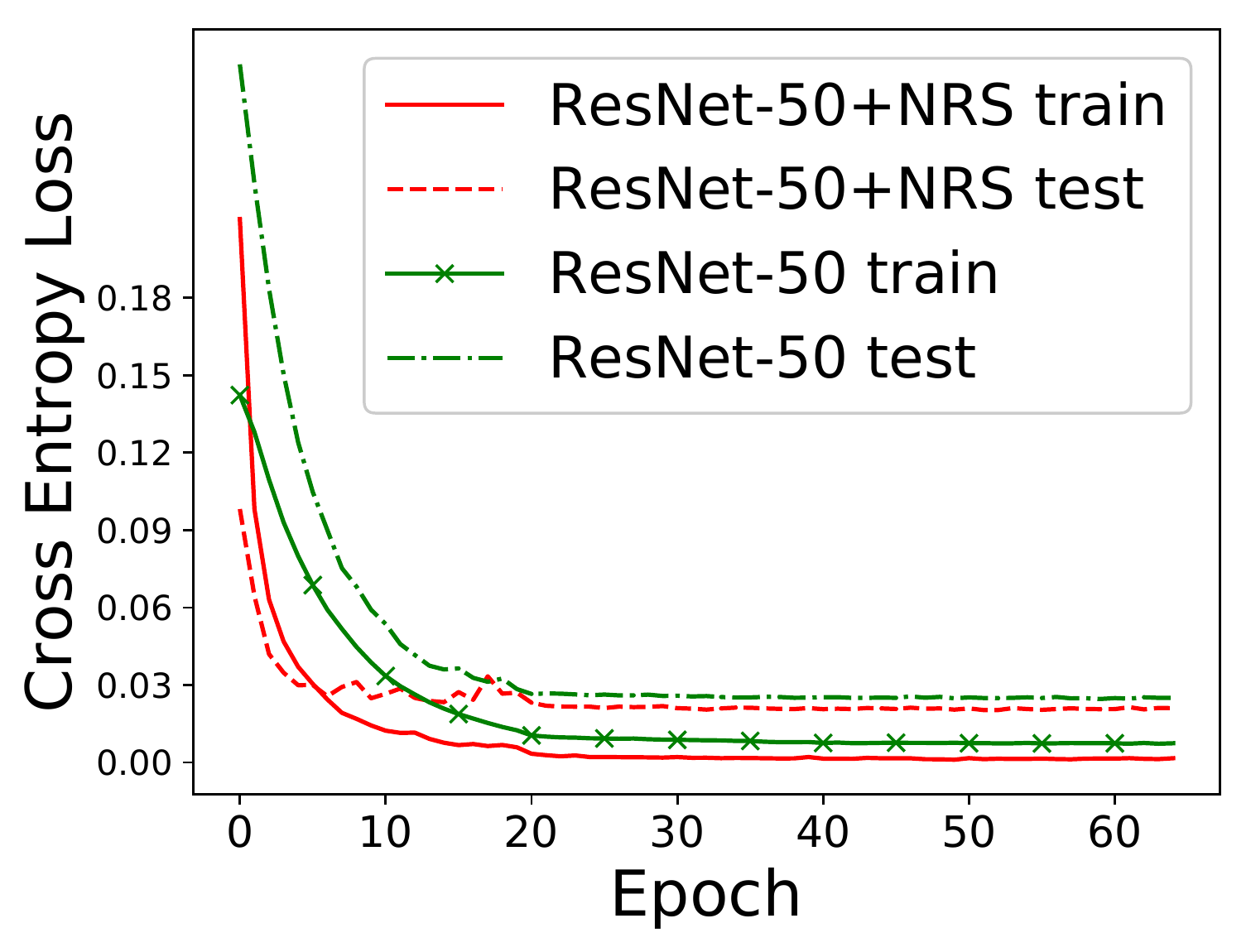}
    \includegraphics[width=0.47\linewidth]{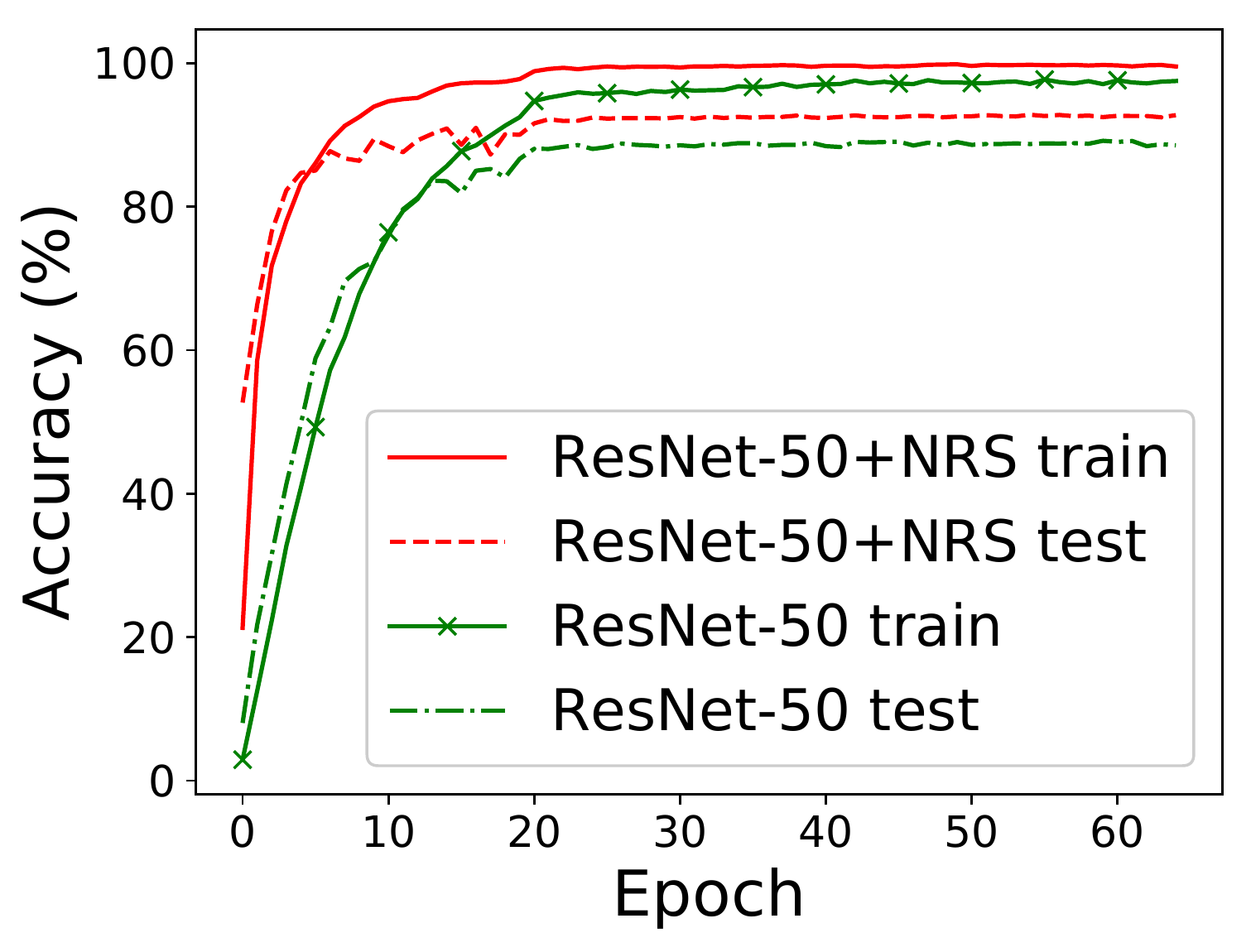}
    \caption{Loss and accuracy learning curves on Aircrafts. Both ResNet-50 and ResNet-50+NFL are trained under the same setting.}
    \label{fig:my_label}
\end{figure}

\subsubsection{ImageNet ILSVRC-12}\label{sec:ImageNet}
We then evaluate NRS on the large-scale ImageNet ILSVRC-12 task and also NRS is used after GAP at the end of the network.

\textbf{Implementation details:} We train a ResNet-50+NRS model from scratch on ImageNet, which is described in Sec~\ref{sec:CUB200-2011} except that the last layer is a 1000-way softmax layer. The images are resized with shorter side=256, then a $224\times224$ crop is randomly sampled from the resized image with horizontal flip and mean-std normalization. Then, the preprocessed images are fed into ResNet-50+NRS model. We train ResNet-50+NRS using SGD with batch size of 256, a momentum of 0.9 and a weight decay of 1e-4 for 100 epochs. The initial learning rate starts from 0.1, and is devided by 10 every 30 epochs. A ResNet-18+NRS model is constructed and trained in a similar way, except that we set $nMul$ and $nPer$ to 4 and 32, respectively. For MobileNetV2~\cite{mobilenetv2:sabdker:CVPR18}, we set $nMul$ and $nPer$ to 1 and 32, respectively. We train the network using SGD with batch size of 256, a momentum of 0.9 and a weight decay of 4e-5 for 150 epochs. We initialize the learning rate to 0.05 and use cosine learning rate decay.

\textbf{Comparison with baseline methods:} Table~\ref{tab:resnet50-imagenet-1crop} shows that NRS produces 0.70\%, 1.92\% and 0.73\% top-1 error (1-crop) less than the original MobileNetV2, ResNet-18 and ResNet-50 model, respectively, with negligible increase in parameters and FLOPs. It indicates that our NRS method is also effective for large-scale recognition, achieving better performance consistently under various architectures.

\begin{table}[t]
	\caption{Error rate (\%, 1-crop prediction) comparison on ImageNet ILSVRC-12 under different architectures. $\blacktriangle$ denotes our method.}
	\label{tab:resnet50-imagenet-1crop}
	\centering
	\renewcommand{\arraystretch}{0.7}
	\setlength{\tabcolsep}{3.9pt}
	\begin{threeparttable}
		\begin{tabular}{l|c|c|c}
			\hline
			Method    & \#Params& \#FLOPs & Top-1 / Top-5 error \\
			\hline
			Original ResNet-50\tnote{1} & 25.56M & 4135.79M & 23.85 / \phantom{0}7.13\\
			ResNet-50+NRS $\blacktriangle$ & 29.98M & 4142.26M & \textbf{23.12} / \phantom{0}\textbf{6.62} \\
			\hline
			Original ResNet-18\tnote{1} & 11.69M & 1824.55M & 30.24 / 10.92\\
			ResNet-18+NRS $\blacktriangle$ & 13.82M & 1828.22M & \textbf{28.32} / \phantom{0}\textbf{9.77} \\
			\hline
			Original MobileNetV2\tnote{1} & \phantom{0}3.50M & \oo328.77M & 28.12 / \phantom{0}9.71\\
			MobileNetV2+NRS $\blacktriangle$& \phantom{0}3.88M & \oo329.21M & \textbf{27.42} / \phantom{0}\textbf{9.39} \\
			\hline
		\end{tabular}
		\begin{tablenotes}
			\footnotesize
			\item[1] \url{https://pytorch.org/docs/master/torchvision/models.html}
		\end{tablenotes}
	\end{threeparttable}
\end{table}

\subsubsection{NRS across all layers}\label{sec:senet}
Motivated by the Squeeze-and-Excitation (SE) method~\cite{senet:hujie:arxiv}, we use NRS to replace all the SE modules. We conduct experiments on CIFAR-10~\cite{cifar}, CIFAR-100~\cite{cifar} and the ImageNet ILSVRC-12 task. We compare our method with baseline methods and SENet under various architectures.

\begin{figure}[htbp]
	\centering
	\includegraphics[width=\columnwidth]{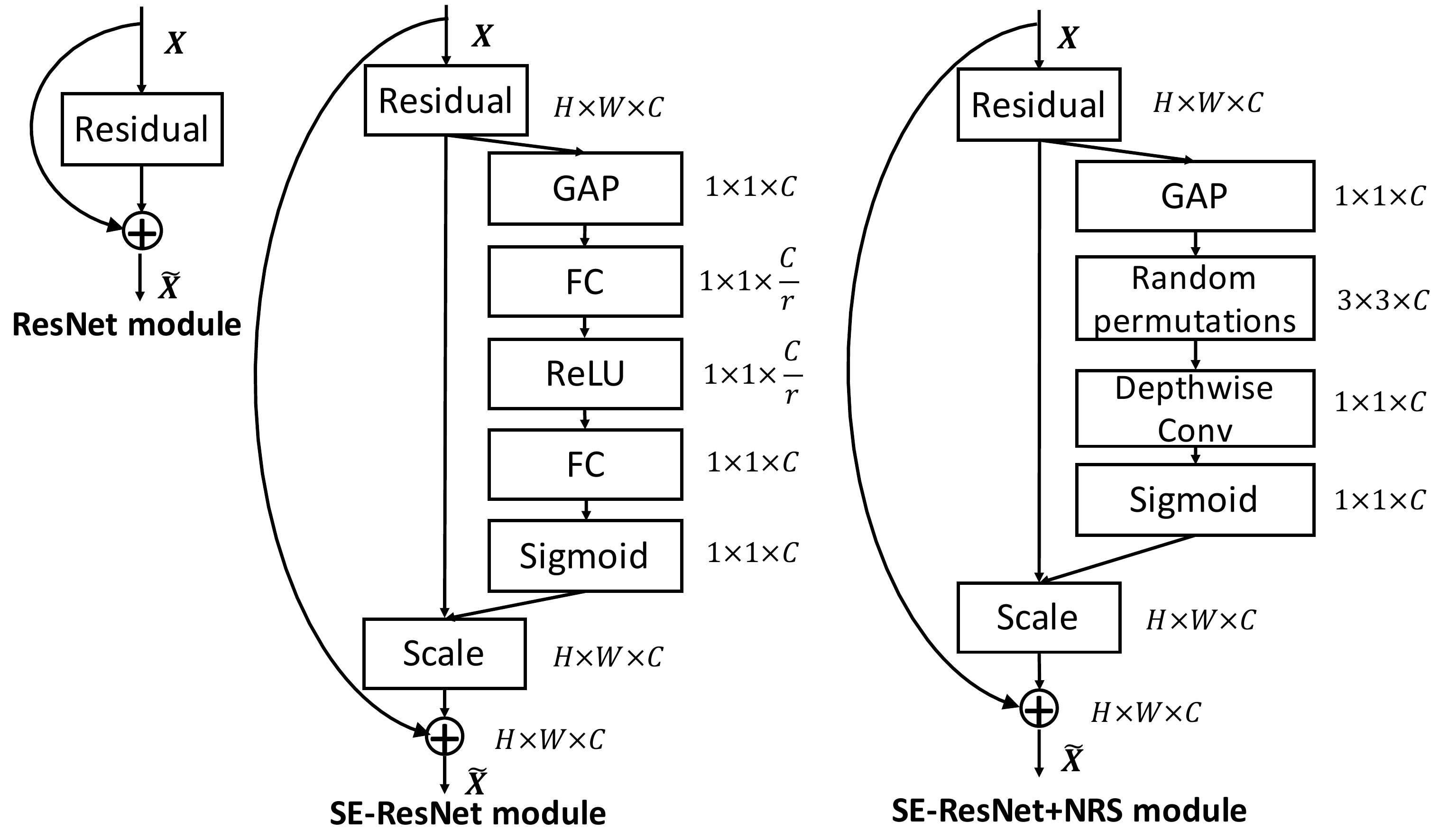}
	\caption{The schema of the original Residual module (left), the SE-ResNet module (middle) and the SE-ResNet+NRS module (right).}
	\label{fig:se-module}
\end{figure}

\begin{table}[t]
	\caption{Comparison of params, FLOPs and accuracy (\%) on CIFAR-10 and CIFAR-100 under various architectures. $\blacktriangle$ denotes our method.}
	\label{tab:cifar-result}
	\centering
	\renewcommand{\arraystretch}{0.7}
	\setlength{\tabcolsep}{1.1pt}
	\renewcommand{\multirowsetup}{\centering}
	\begin{tabular}{l|c|c|c|c|c|c}
		\hline
		\multirow{2}{*}{Method} &\multicolumn{3}{c|}{CIFAR-10} & \multicolumn{3}{c}{CIFAR-100}\\
		\cline{2-7}
		&\#Params &\#FLOPs & Acc. & \#Params & \#FLOPs & Acc. \\
		\hline
		original ResNet-20&\phantom{0}0.27M&\phantom{0}\phantom{0}41.62M&92.75&\phantom{0}0.28M&\phantom{0}\phantom{0}41.63M&69.33\\ 
		SE-ResNet-20&\phantom{0}0.27M&\phantom{0}\phantom{0}41.71M&93.28&\phantom{0}0.28M&\phantom{0}\phantom{0}41.72M&70.35\\ 
		SE-ResNet-20+NRS $\blacktriangle$& \phantom{0}0.28M&\phantom{0}\phantom{0}41.71M &\textbf{93.73}&\phantom{0}0.28M&\phantom{0}\phantom{0}41.72M&\textbf{70.38}\\ 
		\hline
		original ResNet-50&23.52M&1311.59M&95.78&23.71M&1311.96M&80.41\\ 
		SE-ResNet-50&26.04M&1318.42M&95.59&26.22M &1318.79M &\textbf{81.57}\\ 
		SE-ResNet-50+NRS $\blacktriangle$ &23.67M&1313.56M&\textbf{96.05}& 23.86M&1313.93M &81.48\\
		\hline
		original Inception-v3&22.13M&3411.04M&94.83 &22.32M&3411.41M&79.62\\ 
		SE-Inception-v3&23.79M&3416.04M&95.60& 23.97M&3416.41M &80.44\\ 
		SE-Inception-v3+NRS $\blacktriangle$ &22.23M&3412.85M& \textbf{95.67}& 22.42M&3413.22M &\textbf{80.54}\\
		\hline
	\end{tabular}
\end{table}

\begin{table}[t]
	\caption{Comparison of parameters, FLOPs, inference time per image (ms) and error rate (\%, 1-crop prediction) comparison on ImageNet ILSVRC-12 under SENet architectures. The inference time is recorded with batch size of 1 on both CPU and GPU. $\blacktriangle$ denotes our method.}
	\label{tab:se-resnet50-imagenet-1crop}
	\centering
	\small
	\renewcommand{\arraystretch}{0.9}
	\setlength{\tabcolsep}{0.8pt}
	\begin{threeparttable} 
		\begin{tabular}{l|c|c|c|c|c|c}
			\hline
			\multirow{2}{*}{Method}    & \multirow{2}{*}{\#Params}& \multirow{2}{*}{\#FLOPs} & \multicolumn{2}{c|}{Inference Time} & \multicolumn{2}{c}{Top-1 / Top-5 error} \\
			\cline{4-7}
			&&&CPU&GPU&reported in \cite{senet:hujie:arxiv}& our results\\
			\hline
			Original ResNet-50\tnote{1} & 25.56M & 4135.79M &465.39&21.07&24.80 / 7.48&23.85 / 7.13\\
			SE-ResNet-50 & 28.07M & 4146.34M &581.32&35.82&23.29 / 6.62& \textbf{22.68} / \textbf{6.30} \\
			SE-ResNet-50+NRS $\blacktriangle$& 25.71M & 4141.54M &523.97&32.80&- / -& 22.89 / 6.57\\
			\hline
		\end{tabular}
		\begin{tablenotes}
			\footnotesize
			\item[1] \url{https://pytorch.org/docs/master/torchvision/models.html}
		\end{tablenotes}
	\end{threeparttable}
\end{table}

\textbf{Implementation details:} We replace the 2 FC layers in each SE block with NRS, specifically, random permutations and 1 group convolution layer followed by sigmoid activation, which is called SENet+NRS, as shown in Figure~\ref{fig:se-module}. In all our experiments in this section, we set $nMul$ and $nPer$ to 1 and $nH$ to 3 for SENet+NRS and the reduction ratio is set to 16 for SENet as is done in \cite{senet:hujie:arxiv}. For CIFAR-10 and CIFAR-100, we use ResNet-20~\cite{resnet:he:CVPR16}, ResNet-50~\cite{resnet:he:CVPR16} and Inception-v3~\cite{inceptionv3:szegedy:cvpr16} as the backbone network. Mean subtraction, horizontal random flip and $32\times32$ random crops after padding 4 pixels on each side were performed as data preprocessing and augmentation. We train all networks from scratch using SGD with 0.9 momentum, a weight decay of 5e-4 and batch size of 128 for 350 epochs. The initial learning rate starts from 0.1 using cosine learning rate decay. For ImageNet, we follow the same setting as in~\cite{senet:hujie:arxiv}. The images are resized with shorter side=256, then a $224\times224$ crop is randomly sampled from the resized imgae with horizontal flip and mean-std normalization. We use SGD with a momentum of 0.9, a weight decay of 1e-4, and batch size of 256 and the initial learning rate is set to 0.15 and decreased by a factor of 10 every 30 epochs. Models are trained for 100 epochs from scratch.

\textbf{Computational cost:} As can be seen in Figure~\ref{fig:se-module}, the SE module has a total number of parameters as $\frac{2C^2}{r}$ and computational cost of $\frac{2C^2}{r}$. While for NRS, according to Equation~\ref{eq:nrs-parameters} and Equation~\ref{eq:nrs-cost}, NRS has both $9C$ parameters and computational cost ($K=3$ here). Thus, we have:
\begin{equation}
    9C \leq \frac{2C^2}{r} \Longrightarrow C \geq \frac{9}{2}r.
\end{equation}
In our experiments we set $r=16$ and we know if $C\geq72$, then our NRS module will have less parameters and computational cost than the original SE module and the reduction will get larger as $C$ increases. Take ResNet-50 as an example, $C$ ranges from 64 to 2048 in ResNet-50, so our NRS module will greatly reduce the total number of parameters and computational cost.  

\textbf{Comparison with baseline methods:} Table~\ref{tab:cifar-result} shows that under ResNet-20, SENet+NRS achieves the highest accuracy on CIFAR-10 and CIFAR-100 with negligible increase in parameters and FLOPs. For ResNet-50 and Inception-v3 backbone, SENet+NRS achieves comparable or better accuracy than original SENet despite using fewer parameters and FLOPs, further confirming the effectiveness of NRS. 

Table~\ref{tab:se-resnet50-imagenet-1crop} shows that under ResNet-50, SENet+NRS achieves fewer parameters, FLOPs and real running time than original SENet while maintaining comparable accuracy. SENet+NRS also achieves higher accuracy than the baseline method with negligible increase in parameters and FLOPs. It indicates that NRS can be integrated not only at the end of a CNN as shown in the previous sections but also across all layers in a CNN to learn non-linear mapping effectively.

\subsection{3D Recognition}
\label{sec:3d}
We then move from 2D image recognition to 3D recognition in this section. 

Common types of 3D objects/scenes include point clouds, polygonal meshes, volumetric grids and multiple view images. 
Among these 3D representations, we are particularly interested to utilize point clouds for 3D understanding tasks, because of two reasons. First, a point cloud is the closest representation to raw collected sensor data. It encodes full information from sensors, without any quantization loss (volumetric grids) or projection loss (multi-view images). Second, a point cloud is quite neat in form, just a collection of points, which avoids the combinatorial irregularities and complexities compared with meshes (e.g. choices on the type, size and connectivity of polygons). The point cloud is also free from the necessity to choose resolution as in volumetric representations, or projection viewpoint in multi-view images.


For point cloud based 3D object recognition tasks, NRS is used after the global feature encoders and we evaluate its effectiveness on ModelNet40~\cite{ModelNet40} under PointNet~\cite{pointnet}, PointNet++~\cite{pointnet++} and DGCNN\cite{dgcnn}. We compare baselines enhanced by NRS with their counterparts and most recent works, i.e., PointConv\cite{pointconv} and RSCNN\cite{rscnn}.

\textbf{Implementation details:} For PointNet\cite{pointnet} and PointNet++\cite{pointnet++}, we train them using Adam\cite{adam} optimizer with L2 regularisation. The initial learning rate is set to be 1e-3, and is decayed by a factor of 0.7 for every 20 epochs. We train the model for 200 epochs. We use a batch size of 24 and the momentum in batch normalisation is 0.9. We represent each object with 1024 points, for PointNet++ we also utilise the surface normal of each point. For the NRS modules, we set $nMul$, $nPer$, and $nH$ to 1, 32 and 3, respectively and the training protocols remain the same as baseline models. During training process, we randomly drop, rescale and translate the point cloud for the augmentation.

\textbf{Comparison with baseline methods}: Table~\ref{tab:3d} shows that both PointNet+NRS and PointNet+++NRS achieve higher accuracy than the corresponding baseline methods with negligible increase in parameters, FLOPs and inference time. When compared with the state-of-the-art method RSCNN~\cite{rscnn}, we get comparable result by integrating our NRS into DGCNN~\cite{dgcnn} (93.1 v.s. 93.6). It is worth noting that the results show that our NRS has the potential to further boost performance when combined with emerging state-of-the-art algorithms, e.g., RSCNN. We prove that NRS can be integrated within the CNN models that working not only on 2D images, but also 3D objects efficiently and effectively.
\begin{table}[t]
	\caption{Comparison of parameters, FLOPs, inference time per point cloud objects (ms) and accuracy (\%) comparison on ModelNet40 under various architectures. The inference time is recorded with batch size of 1 on both CPU and GPU. $\blacktriangle$ denotes our method. `N/A' means that RSCNN doesn't provide a CPU implementation.}
	\label{tab:3d}
	\centering
	\small
	\renewcommand{\arraystretch}{0.8}
	\setlength{\tabcolsep}{0.8pt}
	\begin{threeparttable} 
		\begin{tabular}{l|c|c|c|c|c}
			\hline
			\multirow{2}{*}{Method}    & \multirow{2}{*}{\,\#Params\,}& \multirow{2}{*}{\,\#FLOPs\,} & \multicolumn{2}{c|}{\,\,Inference Time \,} & \multirow{2}{*}{\,Accuracy\,} \\\cline{4-5}
			& & & \quad CPU \,\,\quad & GPU & \\\hline
			PointNet~\cite{pointnet} & \oo3.47M & \oo446.65M & \oo\oo27.06 & \oo\oo7.18 & 89.2 \\
			PointNet+NRS $\blacktriangle$ & \oo3.77M & \oo446.94M & \oo\oo28.69 & \oo\oo7.25 & 90.3 \\
			\hline
			PointNet++~\cite{pointnet++} & \oo1.48M & \oo866.53M & \oo234.16 & 248.03 & 91.9 \\
			PointNet+++NRS $\blacktriangle$ & \oo1.78M & \oo866.83M & \oo235.98 & 248.97 & 92.3 \\
			\hline
			DGCNN~\cite{dgcnn} & \oo1.81M & 2495.23M & 3771.61 & 353.52 & 92.9 \\
			DGCNN+NRS $\blacktriangle$ & \oo2.99M & 2496.42M & 3878.64 & 359.98 & 93.1 \\
			\hline
			PointConv~\cite{pointconv} & 19.57 M & 1461.74 M & \oo270.12 & 237.48 & 92.5 \\\hline
			RSCNN~\cite{rscnn} & \oo1.29 M & \oo286.60 M & {\oo}N/A & \oo28.46 & \textbf{93.6} \\\hline
		\end{tabular}
	\end{threeparttable}
\end{table}

\section{Conclusions}
We proposed a deep learning based random subspace method NRS. We introduced the random subspace method into deep learning with random permutations acting as resampling and group convolutions acting as aggregation, where each base learner learns from a random subset of features. NRS can handle vectorized inputs well and can be installed into CNNs seamlessly both at the end of the network and across all layers in the network for both 2D and 3D recognition tasks. On one hand, it enriches random subspaces with the capability of end-to-end representation learning as well as pervasive deep learning software and hardware support. On the other hand, it effectively learns non-linear feature representations in CNNs with negligible increase in parameters, FLOPs and real running time. We have successfully confirmed the effectiveness of NRS on standard machine learning datasets, popular CIFAR datasets, challenging fine-grained benchmarks, the large-scale ImageNet dataset as well as 3D recognition dataset ModelNet40. In the future, we will continue exploration on combining deep learning and traditional ensemble learning algorithms to better understand the relation between different approaches. Furthermore, we will extend NRS to handle datasets with high dimensional sparse features, as well as small datasets.

%
%
\bibliographystyle{splncs04}
\bibliography{main}


\end{document}